\pdfoutput=1

\documentclass[12pt,oneside]{CUNY_PhD}

\title{Finite Gaussian Neurons:\\
Defending against adversarial attacks \\
by making neural networks say ``I don't know"}
\author{Felix Grezes}

\usepackage{graphicx}
\usepackage{amssymb}
\usepackage{url}
\usepackage{epstopdf}
\usepackage{enumitem}
\usepackage{amsmath}
\usepackage{subcaption}
\usepackage{booktabs}
\usepackage{natbib}
\setcitestyle{authoryear,open={(},close={)}} 

\usepackage[usenames,dvipsnames]{color}
\usepackage[colorlinks=true]{hyperref}
\hypersetup{
  linkcolor      = {RoyalBlue},
  citecolor      = {RoyalBlue},
  urlcolor       = {RoyalBlue}}

\begin{document}

\frontmatter

\maketitle 

\makecopyrightpage

\makeapprovalpage{Pr.\ Michael I.\ Mandel}{Pr.\ Rivka Levitan}{Pr.\ Ioannis Stamos}{Dr.\ Andrew Rosenberg}{Pr.\ Ping Ji}

\makeabstractpage{Pr. Michael I. Mandel}{
Since 2014, artificial neural networks have been known to be vulnerable to adversarial attacks, which can fool the network into producing wrong or nonsensical outputs by making humanly imperceptible alterations to inputs.

While defenses against adversarial attacks have been proposed, they usually involve retraining a new neural network from scratch, a costly task.

In this work, I introduce the Finite Gaussian Neuron (FGN), a novel neuron architecture for artificial neural networks\protect\footnote{The code used for this work is available at \url{https://github.com/grezesf/FGN---Research} under the GPL 3.0 open source license. Work done with PyTorch.}.
\newpage 

My works aims to:
\begin{itemize}[itemsep=-4mm]
    \item easily convert existing models to Finite Gaussian Neuron architecture,
    \item while preserving the existing model's behavior on real data,
    \item and offering resistance against adversarial attacks.
\end{itemize}

I show that converted and retrained Finite Gaussian Neural Networks (FGNN) always have lower confidence (i.e., are not overconfident) in their predictions over randomized and Fast Gradient Sign Method adversarial images when compared to classical neural networks, while maintaining high accuracy and confidence over real MNIST images.
To further validate the capacity of Finite Gaussian Neurons to protect from adversarial attacks, I compare the behavior of FGNs to that of Bayesian Neural Networks  against both randomized and adversarial images, and show how the behavior of the two architectures differs.
Finally I show some limitations of the FGN models by testing them on the more complex SPEECHCOMMANDS task, against the stronger Carlini-Wagner and Projected Gradient Descent adversarial attacks.


}


\tableofcontents
\listoffigures
\mainmatter

\chapter{Introduction}
Artificial neurons, first created as the Threshold Logic Unit by \citet{mcculloch1943logical}, made trainable as the perceptron by \citet{rosenblatt1958perceptron}, and integrated into practical artificial neural networks by \citet{werbos1975beyond},
are a fundamental building block of modern machine learning and artificial intelligence systems. Despite their success, artificial neural networks were shown in 2014  to be vulnerable to adversarial attacks \citep{goodfellow2014explaining}. These adversarial attack methods exploit unintuitive or misunderstood properties of high-dimensional vector spaces to generate adversarial examples, i.e., carefully crafted inputs that are capable of fooling these networks. These adversarial examples are often indistinguishable from normal inputs to humans. 
Defenses against adversarial attacks have been proposed (survey by \citet{akhtar2018threat}), but are either computationally expensive as in the case of adversarial re-training \citep{madry2019deep}, or do not generalize well to novel attacks as in the case of network distillation \citep{Papernot2016DistillationAA,carlini2017evaluating}. See section \ref{rel-work} for further comparison with existing methods. 

In this work I introduce the Finite Gaussian Neuron (FGN), a novel artificial neuron architecture that combines the classical artificial neuron with an Gaussian component to restrict the neuron's range of activity to a finite area of the input space close to training samples.
I show that artificial neural networks that incorporate the FGN architecture (FGNN) are resistant to adversarial attacks, while exhibiting another desirable property: they are naturally resistant to out of domain inputs. Furthermore, existing networks can be converted to the FGN architecture without any expensive computation while preserving the network's behavior over data.

\newpage
\section{Motivation}
There are two main intuitions that motivate the Finite Gaussian Neuron, both of which might explain why neural networks are susceptible to adversarial attacks: the \emph{piece-wise linearity} of the artificial neurons and the \emph{curse of dimensionality}.

\begin{figure}[!t]
    \centering
    \includegraphics[width=0.9\textwidth]{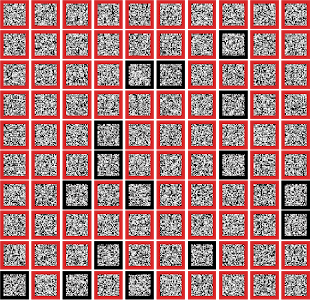}
    \caption[Confident random predictions]{A model trained to recognize MNIST digits (see section \ref{FGNNs over MNIST} for model details) often makes strong predictions over random noise. For each red square, this typical model's softmax output gives over $0.5$ confidence in one of the ten digits, i.e., makes a confident prediction that digit is present in the random image.}
    \label{fig:mot-pred-noise}
\end{figure}

\begin{figure}[!b]
    \centering
    \includegraphics[width=0.9\textwidth]{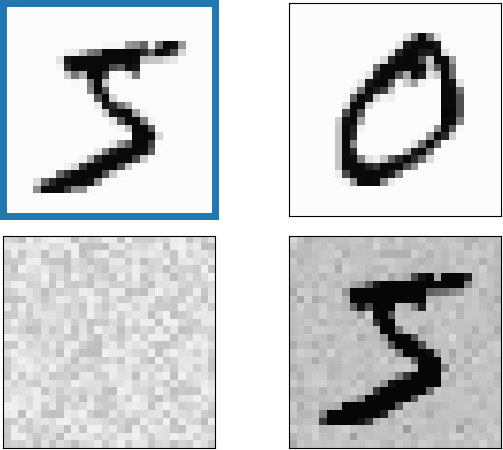}
    \caption[High dimension counter-intuitive behavior]{Three distinct images that all have the same Euclidean distance to the image in the blue box. Human perception doesn't always align with mathematical definitions.}
    \label{fig:mot-dist}
\end{figure}

Typically, artificial neural networks are built by combining artificial neurons into layers, and these neurons individually separate their input space into linear contours, see \citet{goodfellow2016deep}. The combination of these linear contours through stacked layers allow the network to output highly complex and non-linear contours, but a consequence of this linear combination of linear separators is that neural networks tend to have excessive confidence \citep{guo2017calibration} in their output in regions of space far from their training data, see figure \ref{fig:mot-pred-noise}.
The curse of dimensionality refers to many unintuitive phenomena that arise when analyzing data in high-dimensional spaces, see \citep{zimek2012survey}. Notably, distances become hard for humans to visualize. Figure \ref{fig:mot-dist} gives a simplistic example. Other commonly referenced unintuitive phenomena are that high-dimensional spheres have most of their volume concentrated near their surface, and that high-dimensional data sets become easier to linearly separate, see \citep{beyer1999nearest}.

\begin{figure}[!b]
    \centering
    \includegraphics[width=0.9\textwidth]{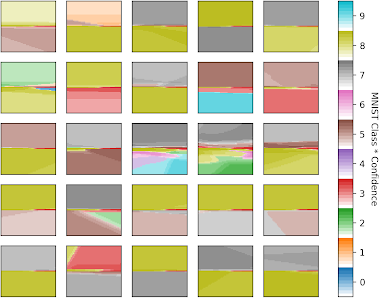}
    \caption[MNIST boundaries around two samples]{What the simple MNIST model from Figure~\ref{fig:mot-pred-noise} predicts when linearly interpolating between two images. Each pixel in these plots shows the class predicted for an interpolated input MNIST image. The centre-leftmost pixel and centre-rightmost pixel correspond to two images pulled from the data in classes 8 (yellow) and 3 (red), respectively. Left-right movement along the center corresponds to linear interpolation of the two images. Each vertical axis corresponds to movement along a different random vector orthogonal to the left-right image-to-image vector. Each color corresponds to an MNIST class prediction. Surprisingly, while the two original images are correctly classified as 8 and 3 respectively, orthogonal movement almost always leads to a different class. An analogy would be: walking in a straight line from Paris to Berlin predictably takes you from France to Germany, but any step left or right of that straight line always lands outside of France or Germany. This image is revisited in Section \ref{Observations}}
    \label{fig:mot-bounds}
\end{figure}

The combination of the piece-wise linearity of neural networks with the various unintuitive curse of dimensionality phenomena lead to various unexpected behaviors of neural networks. For example \citet{szegedy2013intriguing} showed that the boundaries between the various classes in the hyper-space (as predicted by a neural network) are often linear and that most real inputs are close distance-wise to every boundary. Figure \ref{fig:mot-bounds} shows an example of this behavior.
These intuitions motivate the definition of the Finite Gaussian Neuron (in Section \ref{Definition of the FGN}), as a combination of a classical neuron with a Gaussian function of effectively finite support.

\newpage
\section{Related Work}
\label{rel-work}
\subsection{Fast Gradient Sign Method}
\citet{goodfellow2014explaining} introduce adversarial attacks via the Fast Gradient Sign Method (FGSM) which take a fixed size $\epsilon$ step in the \emph{input} space, using gradient descent to find the direction that \emph{maximizes} the network loss $\mathcal{L}$ over the input $x$ with true label $t$ for model parameters $\theta$.
\begin{align}
\textrm{FGSM\_adversary}(x) = x + \epsilon \cdot \textrm{sign}(\nabla_x(\mathcal{L}(\theta,x,t)). \label{eq:fgsm}
\end{align}
Note that this is the opposite direction that is used in training the model and that this is the untargeted version of the FGSM attack.
The authors show that continuous retraining of the model using adversarial examples protects the model from these attacks. This adversarial training method however is computationally expensive as it requires both creating new adversarial examples after each training epoch and continuously retraining the model with a growing dataset. They also show that adversarial training based on the FGSM attack does not necessarily protect against other adversarial attacks.

\subsection{Projected Gradient Descent}
Pushing adversarial training further, \citet{madry2019deep} propose a multi-step variant of FGSM, which is called Projected Gradient Descent (PGD) on the negative loss function. At each step $s$, the adversarial input $a_s$ is projected onto the constraining hyper-sphere using $\mathcal{P}$, which is in itself an optimization problem.
\begin{align}
    x_{s} & = \mathcal{P}_{x,\epsilon}(a_s)\\
     & = \mathcal{P}_{x,\epsilon}(x_{s-1} + \alpha \cdot \textrm{sign}(\nabla_x(\mathcal{L}(\theta,x,t))) \label{eq:pgd}
\end{align}
In this case the constraint is the $\ell^{\infty}$ $\epsilon$-bounded hyper-sphere around the input $x$, making the projection simple to compute, i.e. clip the values along each of the $i$ dimensions: 
\begin{align}
    x_{s,i} = \min(x_i+\epsilon, \max(x_i-\epsilon, a_{s,i}))
\end{align}
The step size $\alpha$ is a tunable parameter of the attack, other parameters are the same as for FGSM in equation \ref{eq:fgsm}.

By showing that many PGD attacks, starting from random points near real inputs, end up in similar local maxima, the authors argue that PGD is the strongest possible adversarial attack that relies on first order information (information from the first derivative of the loss, i.e., the gradients). Neural network models retrained on these PGD adversarial examples are shown to be robust to a variety of other first-order attacks. However these new adversarial examples create more complex boundaries which require larger networks to fit, making PGD-retraining a computationally expensive defense. In their experiments,the authors increased the size of the networks 10-fold, and used between 7 and 100 steps in the PGD attacks.

\subsection{Distillation}
Another proposed defense against adversarial attacks is Distillation, proposed by \citet{Papernot2016DistillationAA}. Originally designed by \citet{hinton2015distilling} as a method to reduce the size of deep neural networks by transferring knowledge from larger to smaller networks, distillation works by training a large network, then using the output prediction vectors as soft-labels to train a smaller network, encoding class similarities in the soft-labels. Instead of aiming to reduce model size, Papernot et al.~show that retraining the same model on the soft-labels provides defense against adversarial attacks by making the network less sensitive to small changes over the input, and requiring a higher average minimum
number of features to be modified in order to create adversarial examples. Similarly to adversarial retraining, distillation requires additional computation to generate the soft-labels and retrain the model.

\subsection{Radial Basis Function Networks}
Radial Basis Function (RBF) networks have also been explored as a more intrinsic defense against out of distribution data and adversarial attacks \citep{moody1989fast, chenou2019radial,zadeh2018deeprbf} and in fact are similar in many ways to the FGNNs. Both change the neuron's architecture to limit its activity to a finite area of its input space.
The output $y$ of an RBF neuron is usually defined as:
\begin{align}
    y &= \sum_{i}^{N} w_i \rho \left(\lVert x-c_i \lVert \right) \label{eq:rbf}
\end{align}
with $x$ the input vector, $N$ the number of dimensions of the input, $c_i$ the centers attached to the network, and $\rho$ the radial basis function, chosen such that $\lim_{\parallel x \parallel \rightarrow \infty} \rho(x) = 0$. The Euclidean norm is usually used. 
In contrast to FGNs, RBFs cannot be created from an existing artificial neuron to mimic the activity of that neuron (see section \ref{conversion}).
RBFs have not become as popular as other deep neural networks techniques, perhaps due to their complexity. The RBF architecture doesn't easily generalize to multiple layers, and requires some pre-processing work to compute the centers based on the available data.  .

\subsection{Bayesian Neural Networks}
Bayesian Neural Networks (BNN) introduced by \citet{mackay1992practical, mackay1992bayesian} attach a probability distribution to each weight and bias of the network and combine them as an ensemble to avoid overconfidence that might be caused by a particular selection of model parameters. BNNs estimate the probability distribution over the weights $w$ of the network that maximize the posterior observations $\mathcal{D}={t_1, t_2, \ldots t_N}$
\begin{align}
    p(w|\mathcal{D},\alpha, \beta)  &\propto p(w|\alpha)p(\mathcal{D}|w,\beta^{-1}) \\
    p(w|\alpha) &= \mathcal{N}(w|0,\alpha^{-1}I)\\
    p(D|w,\beta) &= \prod_{n=1}^N  \mathcal{N}(t_n|y(x_n,w),\beta^{-1}) 
\end{align}
with $\alpha,\beta$ the variance factors for the $\mathcal{N}$ Gaussian functions for the weights and observations respectively.
\citet{bishop2006pattern} explain how
variational inference can be applied to BNNs to efficiently compute the weight distribution estimation  \citep{hinton1993keeping, barber1998ensemble}.
BNNs have the capacity of making no predictions when presented with random noise or inputs unrelated to the training data \citep{jospin2022hands}. BNNs have shown some resistance to adversarial attacks \citep{uchendu2021robustness}. Comparison to FGNs is provided in section \ref{Bayesian}.

\chapter{The Finite Gaussian Neuron}
\section{The Classical Artificial Neuron}
\begin{figure}[!h]
    \centering
    \includegraphics[width=0.9\textwidth]{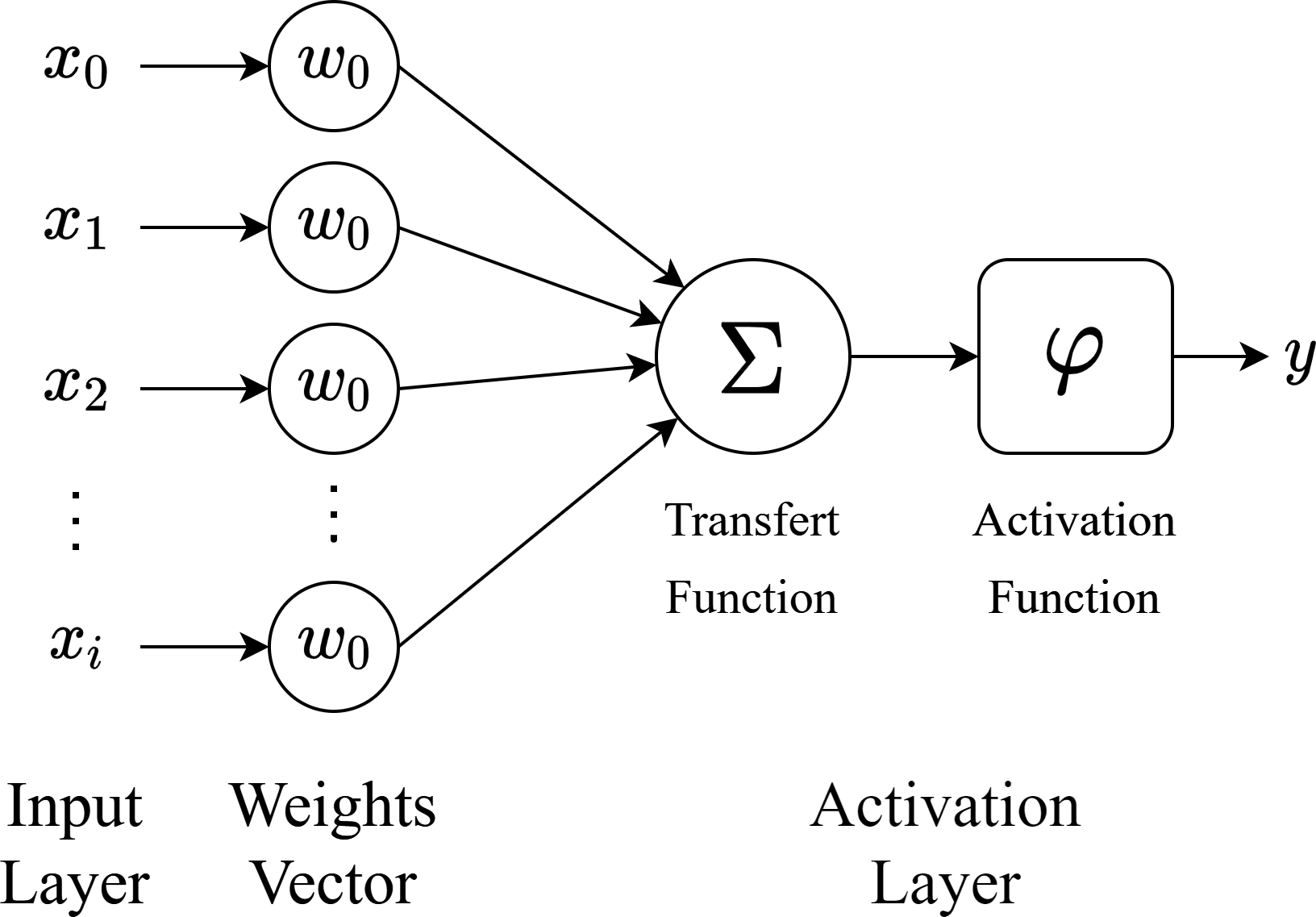}
    \caption{A representation of the classical artificial neuron.}
    \label{fig:classic-neuron}
\end{figure}

A classical artificial neuron's output $y_c$ is defined by: 
\begin{align}
y_{c} &= \varphi(\ell) \label{eq:varphi} \\
\ell &= \sum_{i}w_i \cdot x_i \label{eq:ell}
\end{align}
with $\ell$ being the linear component defined by a linear combination of the inputs $x_i$ and associated weights $w_i$, and with $\varphi$ being the non-linear activation function required by the universal approximator theorem by \citet{cybenko1989approximation, hornik1989multilayer}. The bias term can be implicitly included as an extra input with value 1 or explicitly included. We use the implicit bias representation because it is cleaner. Figure \ref{fig:classic-neuron} gives a visualization of this classical neuron. 

\section{The Finite Gaussian Neuron}
\label{Definition of the FGN}
\begin{figure}[!t]
    \centering
    \includegraphics[width=0.9\textwidth]{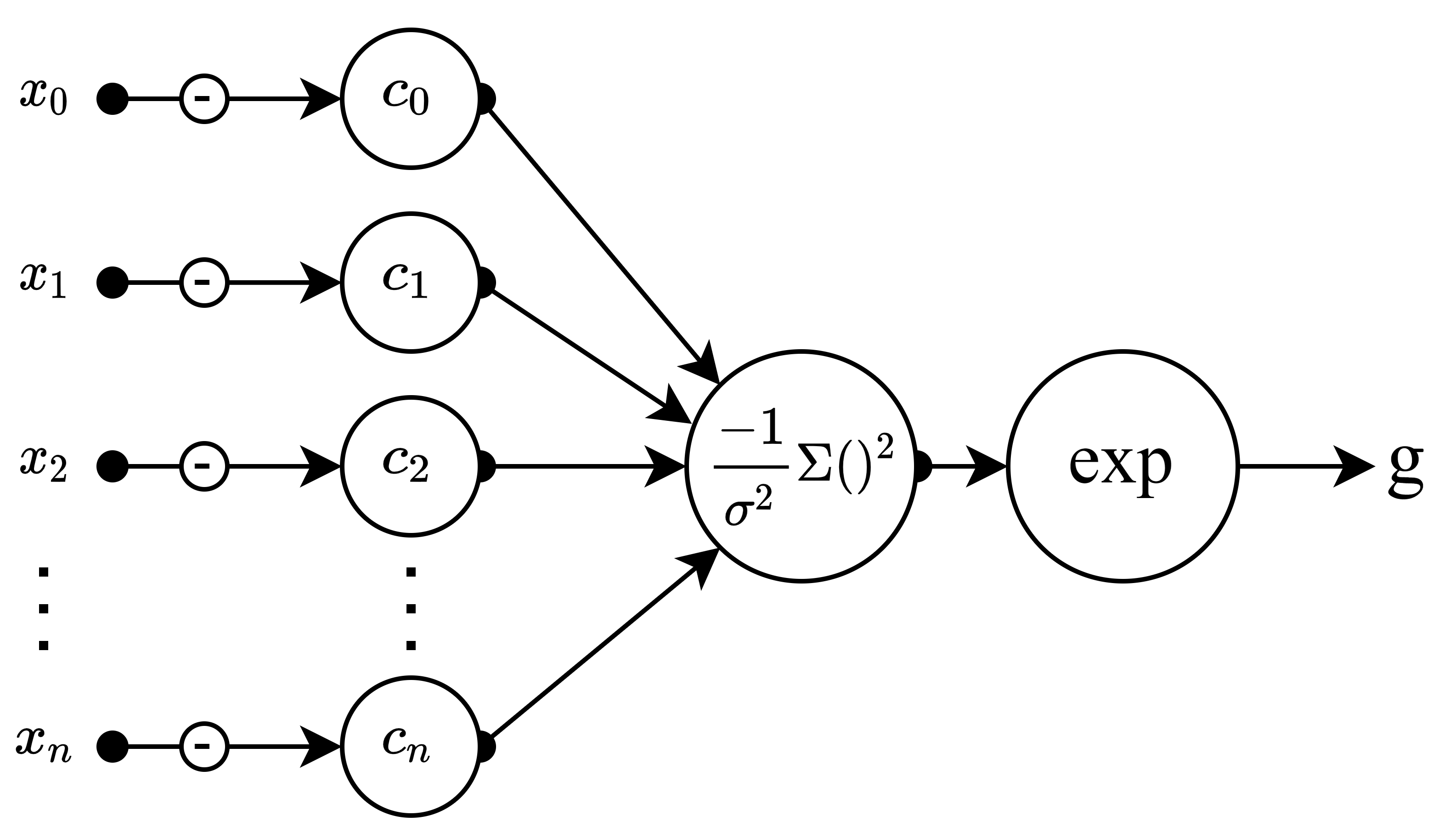}
    \caption{The Gaussian component of a Finite Gaussian Neuron}
    \label{fig:gaussian-comp}
\end{figure}

I now define the Finite Gaussian Neuron (FGN) as a combination of the classical neuron's activity with a Gaussian activity that limits the effective support of the neuron. 
The desired outcome of the FGN definition is to restrict the activity of the neural network to regions of the input hyperspace where data has been observed during training, while making no predictions over regions never observed; with the overarching goal of building FGN networks that are more resistant to adversarial attacks compared to classical networks.\\
Explicitly, the FGN's output $y_f$ is given by:
\begin{align}
y_{f} &= \varphi(\ell) \cdot g \label{eq:fgn}\\
g &= \exp \left(-\frac{1}{\sigma^2}\sum_{i}(x_i-c_i)^2 \right)
\end{align}
with $\ell$ and $\varphi$ the same as in Equations~\eqref{eq:ell} and \eqref{eq:varphi}, i.e., the linear component and non-linear activation function respectively; and with $g$ the new Gaussian component, defined by a center with coordinates $c_i$ that position the neuron in the input hyperspace, and variance $\sigma$ that prevents the neuron's activity from covering the entire input space. If inputs are far away from the center, relative to the variance $\sigma$, then the Gaussian component $g$ approaches value zero and the FGN's output activity will approach zero as well, thus limiting the effective support of the neuron to a limited zone of the input hyperspace. Figure \ref{fig:gaussian-comp} gives a visualization of the new Gaussian component $g$.
The following figures (\ref{fig:classic-heatmap}, \ref{fig:fgn-heatmap}) show the difference in behavior between the classical neuron architecture and the FGN architecture, for an arbitrary neuron, over a two dimensional input space.
\begin{figure}[!t]
    \centering
    \makebox[\textwidth][c]{
    \includegraphics[width=0.5\textwidth]{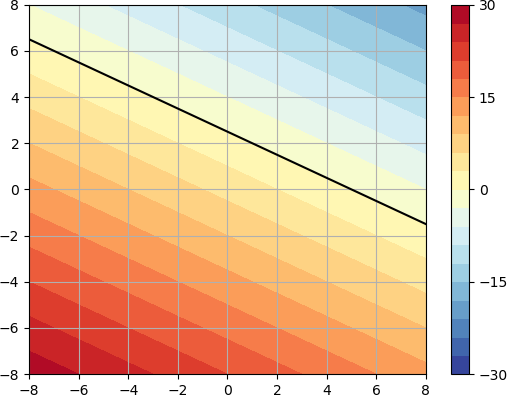}
    \includegraphics[width=0.5\textwidth]{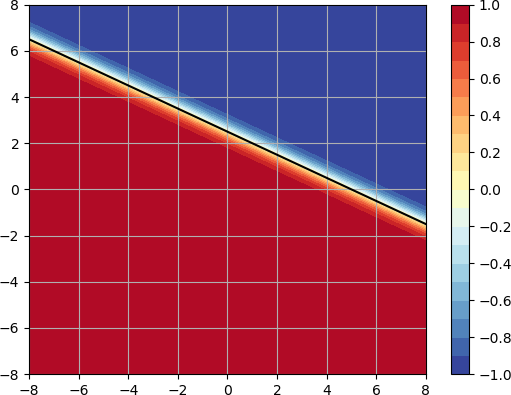}
    }
    \caption[Linear activity map]{On the left, the $\ell = \sum_{i}w_i x_i$ linear component of the classical neuron with weights $W=[-1,-2,5]$ (chosen arbitrarily), shown as an activity heatmap over a 2D input space. On the right the same neuron's output $y_c = \tanh(\ell)$  after passing the linear component through the $tanh$ non-linear activation function. The black line shows where the heatmap value is zero.}
    \label{fig:classic-heatmap}
\end{figure}
\begin{figure}[!t]
    \centering
    \makebox[\textwidth][c]{
    \includegraphics[width=0.5\textwidth]{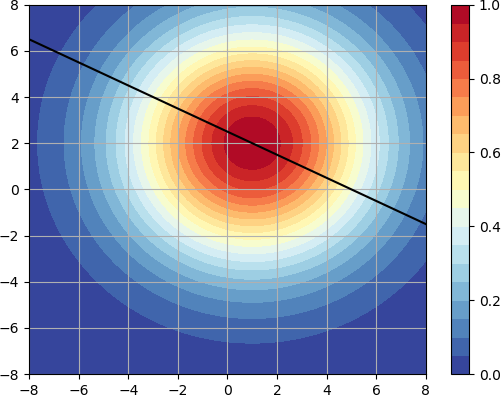}
    \includegraphics[width=0.5\textwidth]{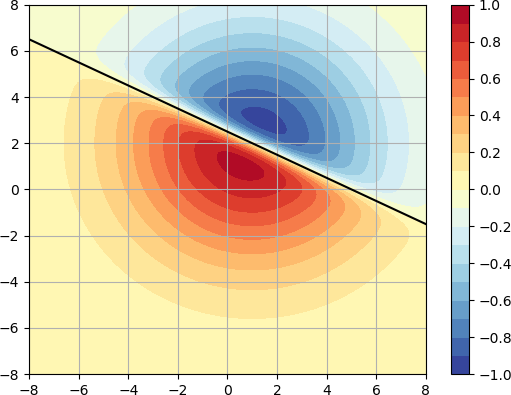}
    }
    \caption[The Gaussian activity map]{On the left, the Gaussian component $g = \exp \left( -\frac{1}{\sigma^2}\sum_{i}(x_i-c_i)^2 \right)$ of the FGN architecture with center $C=[1,2]$ (an arbitrary point on the zero line in the image) and variance $\sigma=5$ (value with both $+1$ and $-1$ activity visible in the image), shown as an activity heatmap over a 2D input space. The line in this image is only for comparison with the other images. On the right, the FGN's output $y_f = \tanh(\ell) \cdot g$ combining the output of the classical neuron with the Gaussian component. Note the limited effective support of the FGN's activity.}
    \label{fig:fgn-heatmap}
\end{figure}

\newpage

\section{Classic Neuron Conversion to FGN}
\label{conversion}
\begin{figure}[!t]
    \centering
    \makebox[\textwidth][c]{
    \includegraphics[width=0.33\textwidth]{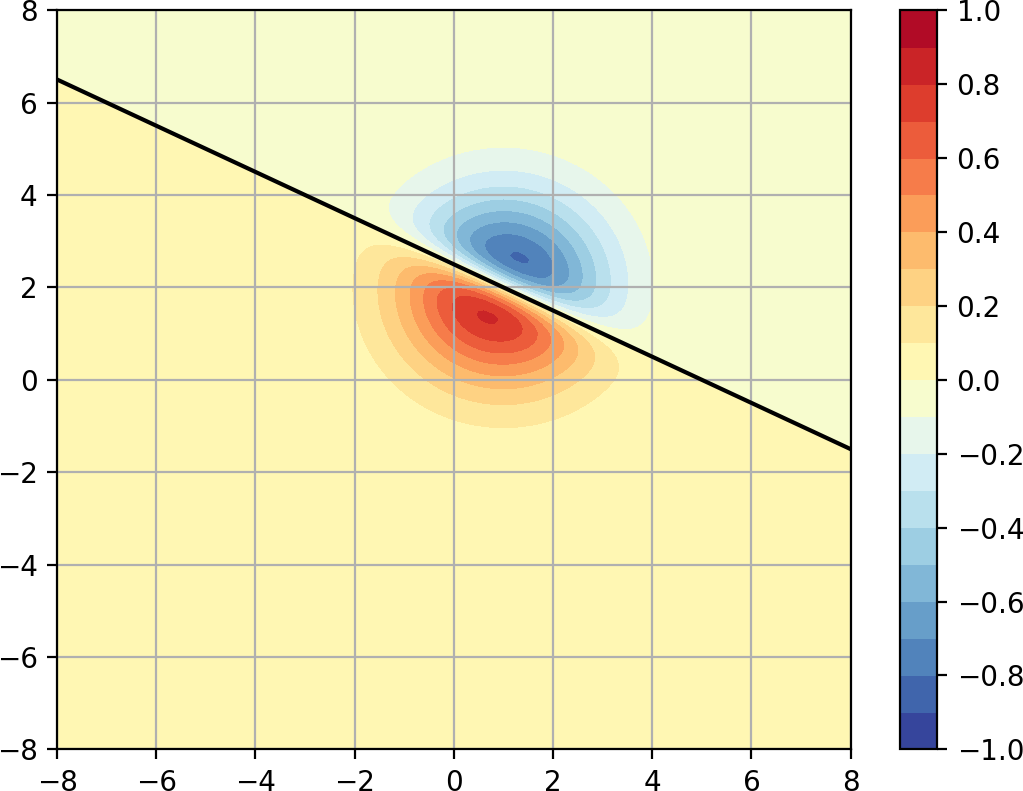}
    \includegraphics[width=0.33\textwidth]{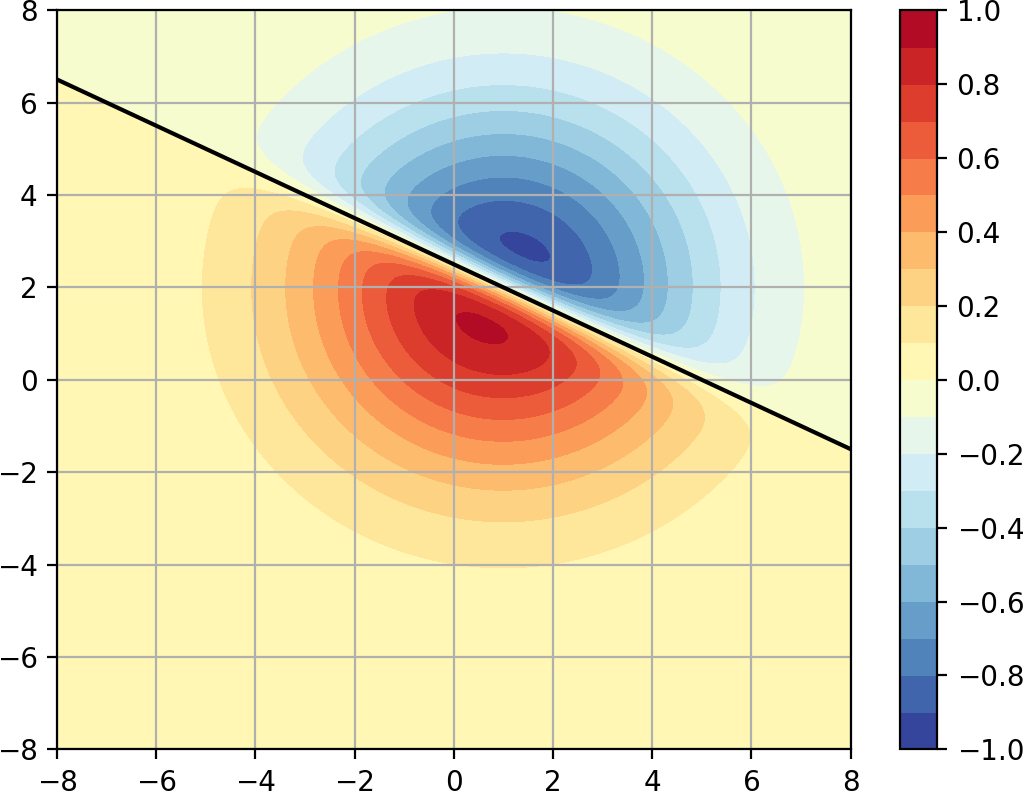}
    \includegraphics[width=0.33\textwidth]{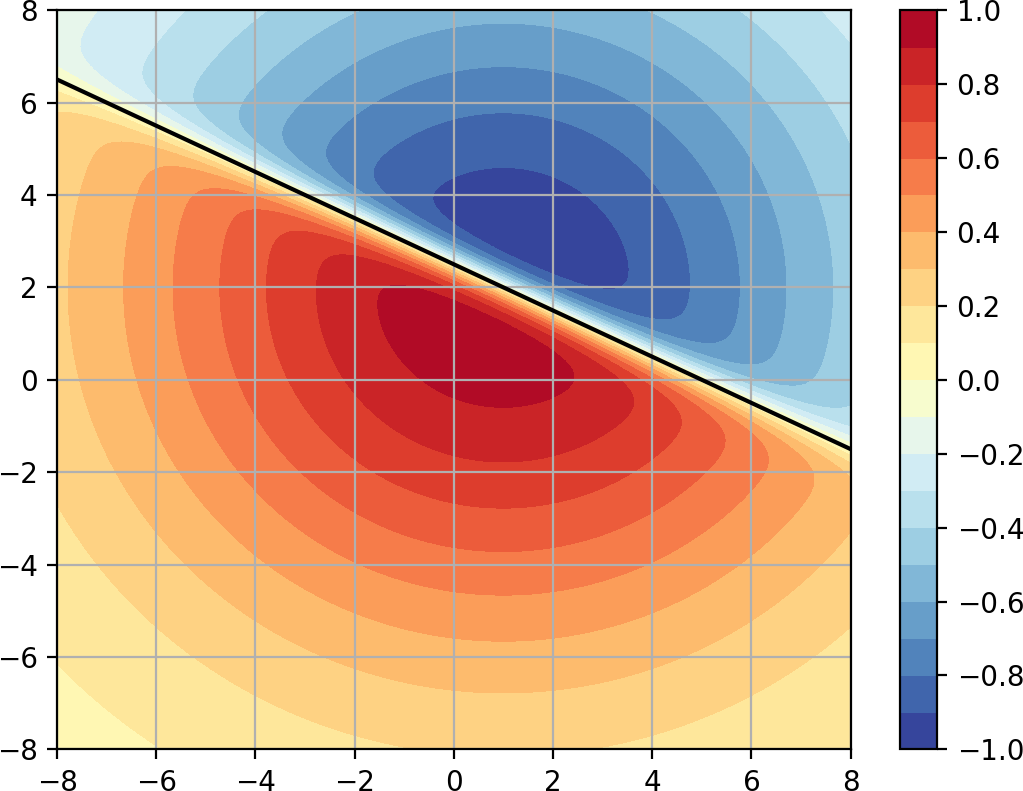}
    }
    \makebox[\textwidth][c]{    \includegraphics[width=0.33\textwidth]{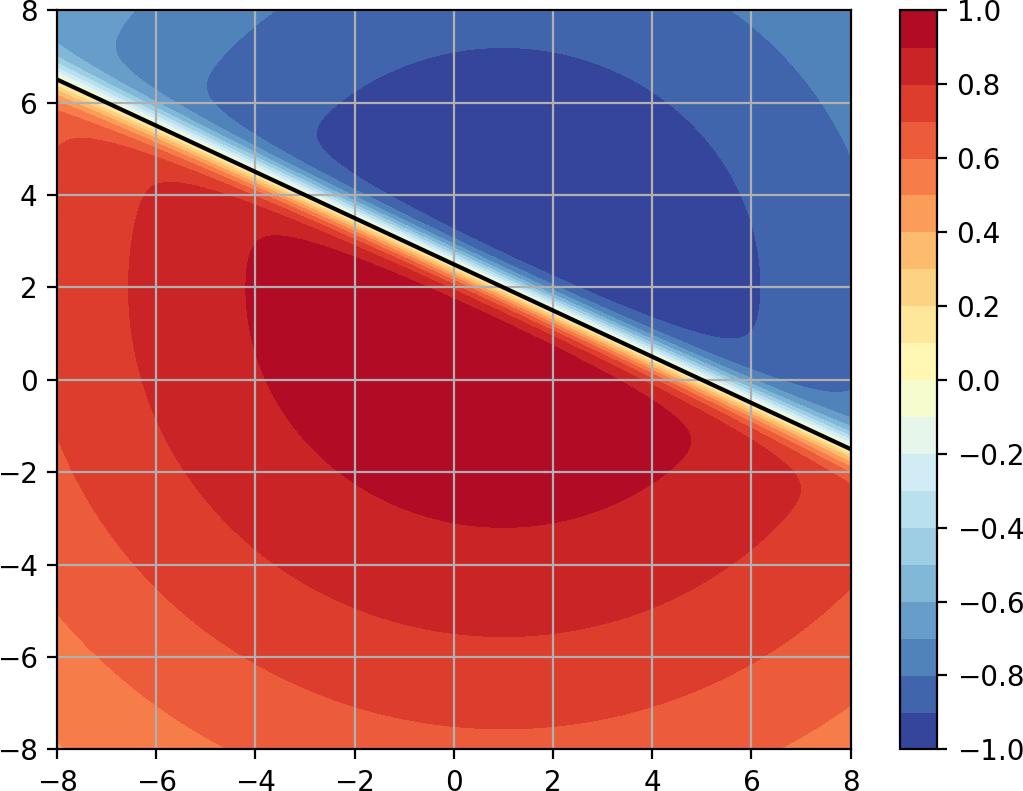}
    \includegraphics[width=0.33\textwidth]{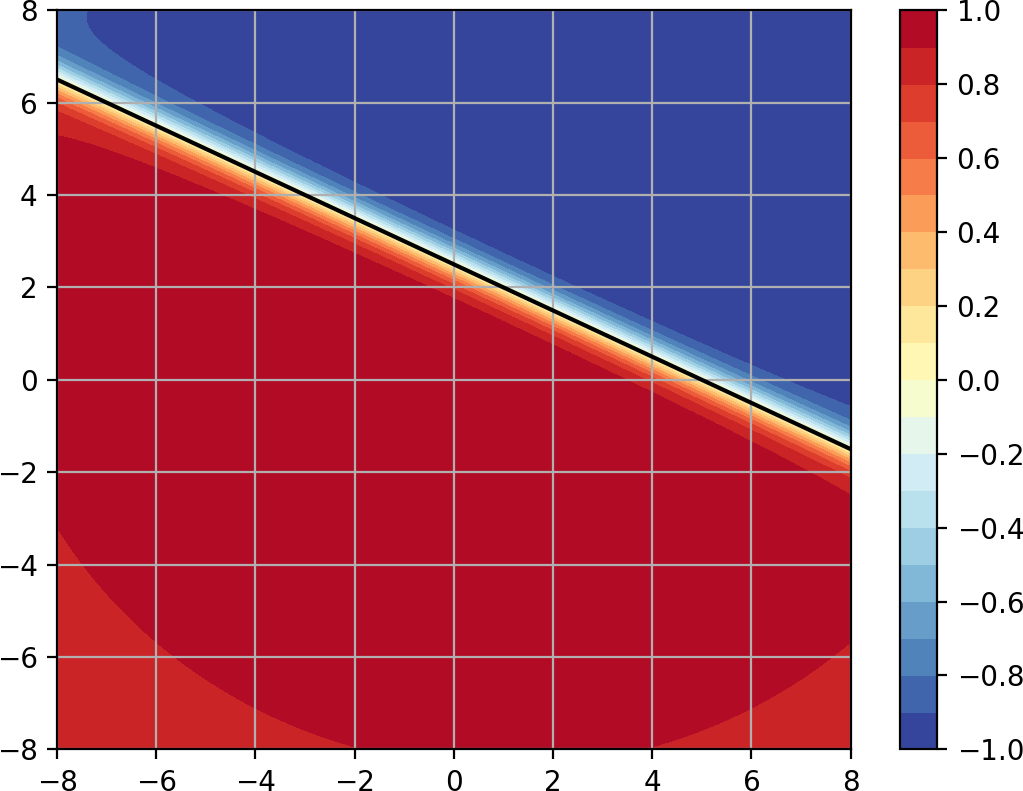}
    \includegraphics[width=0.33\textwidth]{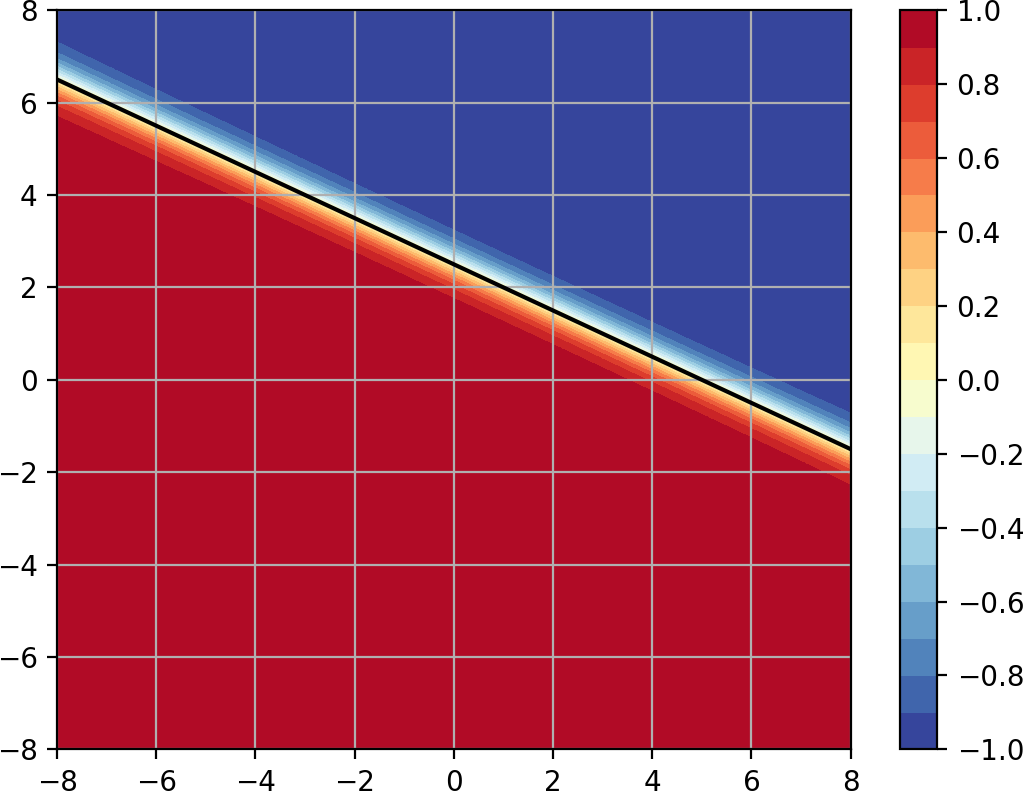}
    }
    \caption[Increasing FGN variance to match classical neuron]{Illustration showing that increasing an FGN's variance $\sigma$ (smallest top-left, largest bottom-right) leads to behavior identical to that of classical neuron with the same weights $W$. Note how the final picture is indistinguishable that shown in figure \ref{fig:classic-heatmap}. }
    \label{fig:matching}
\end{figure}

One important property of the FGN is that existing networks using the classical neuron can be converted to the FGN architecture without changing the network's behavior over a given dataset, and without heavy computation. This is done by converting each classical neuron in the original network to an FGN with identical weight vector $W$ and large variance $\sigma$. Figure \ref{fig:matching} illustrates this property.
Converting a classical neuron to an FGN involves defining two new parameters, the variance $\sigma$ and the center $c_i$. For any given dataset, there will be a unique $C=[c_i]$ that allows for minimum valued variance while not changing the behavior over this dataset. Finding this unique center is the topic of future research. Currently, conversion is done by setting the center $C$ to be the point on the zero-output line (defined by the neuron weights $W$) closest to the origin, and empirically searching for a variance $\sigma$ large enough via unbounded exponential search.

\section{Training the FGN}
Training the FGN is done much the same way as the classical neuron: we find parameters of the FGN that locally minimize a loss function using the back-propagation algorithm. So far, experiments have not shown any specific loss function (Mean-Squared Error, Cross-Entropy, etc\ldots) or gradient descent optimizer algorithm (Adam, RMSprop, etc\ldots) to perform any differently on FGNs versus classical neurons.
Since one of the goals of the FGN is to limit the activity far from the data, we add a regularization term weighted by $\lambda$ to any loss function $\mathcal{L}$ used to train the neuron. This additional term adds pressure to minimize the variance $\sigma$ during training, and $\lambda$ becomes a hyper-parameter to tune to the task like any other.
\begin{align}
    \mathcal{L}  = \tilde{\mathcal{L} } + \lambda\sigma^2. \label{eq:loss}
\end{align}
The partial derivatives of the FGN output $y$ for an input $x_i$  are:
\begin{align}
    \text{Weights:\quad} \frac{\partial y}{\partial w_i} &=  x_i \varphi'(\ell) \cdot g  \\[1em]
    \text{Center:\quad} \frac{\partial y}{\partial c_i} &= \varphi(\ell) \cdot \frac{2(x_i-c_i)}{\sigma^2} \cdot g \\[1em]
    \text{Sigma:\quad} \frac{\partial y}{\partial \sigma} &= \varphi(\ell) \cdot \frac{2\sum_{i}(x_i-c_i)^2}{\sigma^3}\cdot g.
\end{align}
The changes to each of the parameters all depend in part on the Gaussian component $g$. In particular they all need it to be non-zero or they will not change. This shows that the FGN can only learn when the input is close to the neuron centers $c_i$ relative to the variance $\sigma$. Proper initialization is thus important to ensure that the FGN variance and centers cover the data, else the gradients will be non-existent. A simple initialization scheme, used in this work, is to empirically set the variances $\sigma$ to be large enough such that each FGN has some activity over the training data.

As a sanity check, We verify that a single FGN is able to be trained to properly classify a two dimensional linearly separable toy dataset, shown in figure \ref{fig:single-fgn-1}. The following figures \ref{fig:single-fgn-2}, \ref{fig:single-fgn-3} show the FGN's weights $W$, center $c_i$ and variance $\sigma$ adapting to fit the data.
\begin{figure}[!h]
    \centering
    \hspace{0.0\textwidth}
    \makebox[\textwidth][c]{
    \includegraphics[width=0.4525\textwidth]{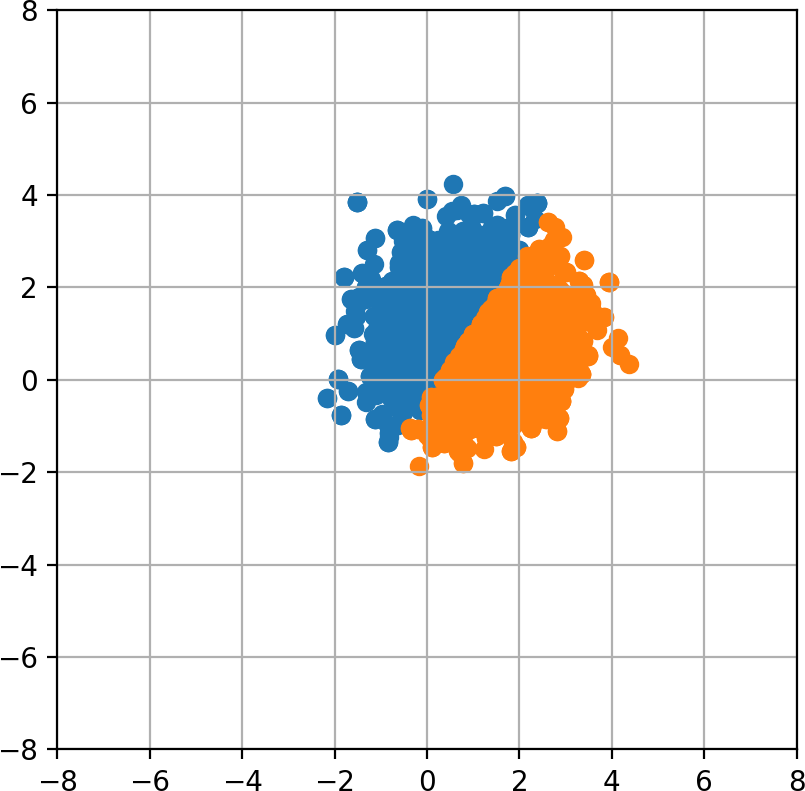}
    \includegraphics[width=0.5475\textwidth]{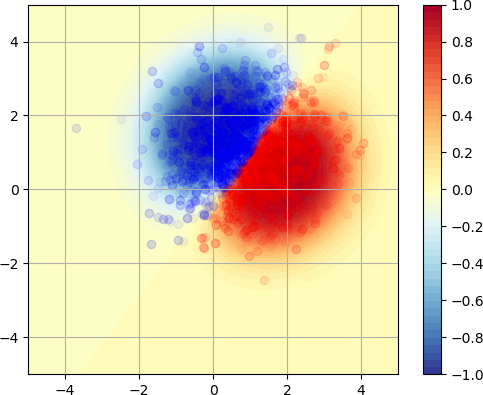}
    }
    \caption[Linearly separable toy 2D data]{The 2D linearly separable toy data centered on $(1,1)$, and the activity of the FGN over the space after training with MSE loss function, $\lambda=0.01$ variance $\sigma$ regularization term, Adam optimizer with $lr=0.05$.}
    \label{fig:single-fgn-1}
\end{figure}
\begin{figure}[!t]
    \centering
    \makebox[\textwidth][c]{
    \includegraphics[width=0.5\textwidth]{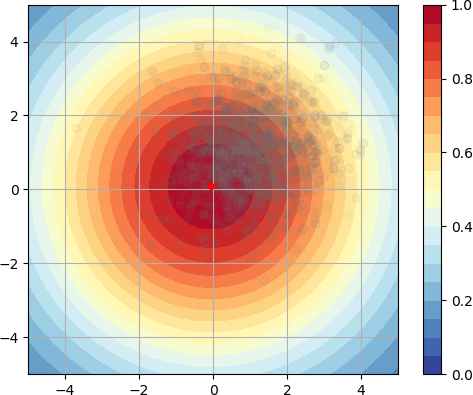}
    \includegraphics[width=0.5\textwidth]{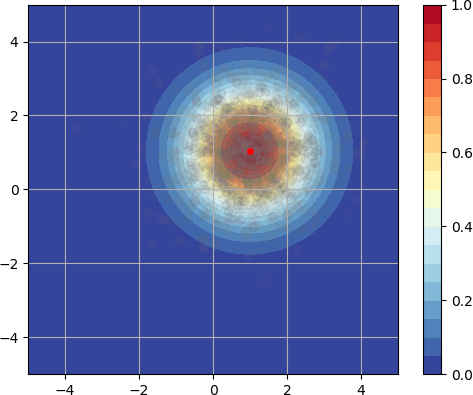}
    }
    \caption[The FGN's Gaussian component activity on toy 2D data]{The FGN's Gaussian component $g$ activity over the 2D space pre and post training. Initially centered on the origin $(0,0)$ with variance $\sigma=5$, after training the Gaussian component is centered on the data and the variance has shrunk such that space far from the data has $g=0$ activity.}
    \label{fig:single-fgn-2}
\end{figure}
\begin{figure}[!h]
    \centering
    \makebox[\textwidth][c]{
    \includegraphics[width=0.5\textwidth,height=6.52cm]{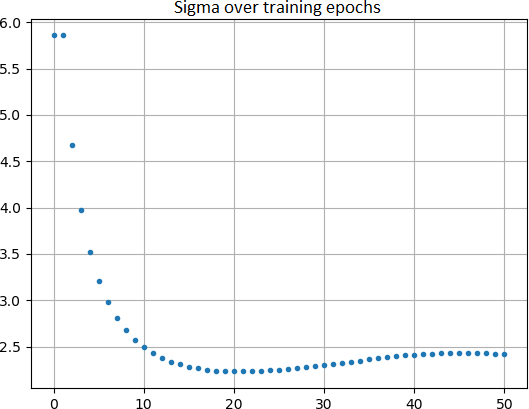}
    \includegraphics[width=0.5\textwidth]{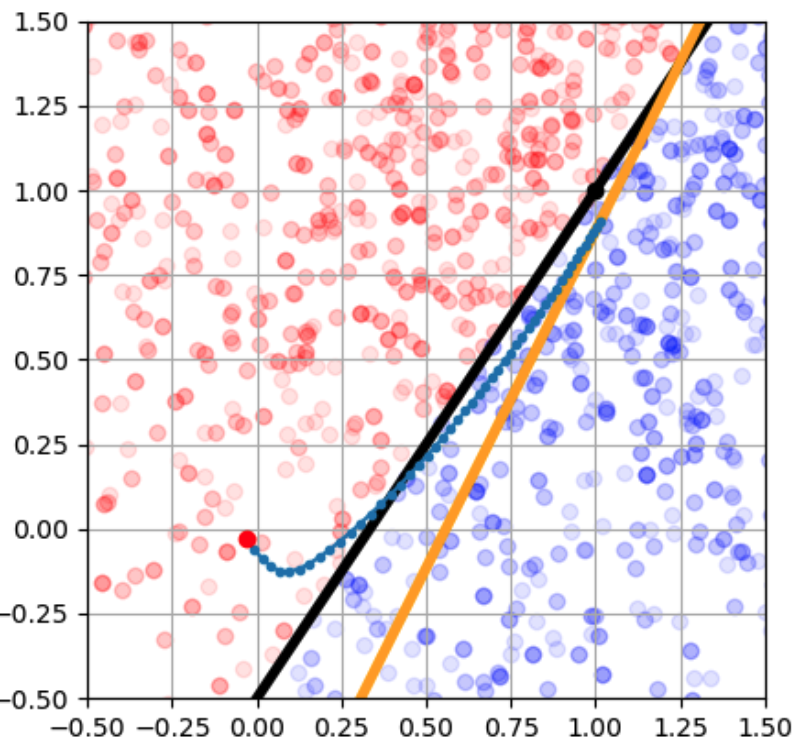}
    }
    \caption[Evolution of the FGN's variance and center parameters]{Evolution of the FGN's variance $\sigma$ and center position during training. On the left, pressured by the loss function's regularizer term $\lambda$,  the variance $\sigma$ shrinks as much as possible while still fitting the data.  On the right, the dotted blue line shows the FGN center's path during training, starting from the red dot in $(0,0)$ and moving towards the theoretically optimal center $(1,1)$ (black dot). The black line is the class border and the orange line is the FGN's predicted border, given by the weights $W$.}
    \label{fig:single-fgn-3}
\end{figure}

\newpage

\section{Variants}
\begin{figure}[!t]
    \centering
    \includegraphics[width=0.66\textwidth]{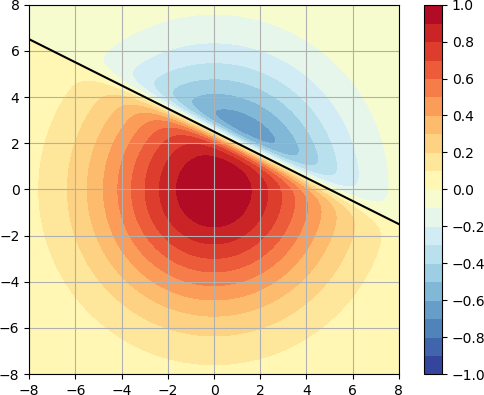}
    \caption{Activity of an FGN with a decoupled bias and centers.}
    \label{fig:decoupled}
\end{figure}

Within the framework of the FGN, there are several modifications that may improve the FGN's performance on specific tasks.
An FGN built to match the behavior of a classical neuron will define the bias term of the linear component $\ell$ by the centers of the FGN, so that the line of zero activity for the linear component passes through the center. This isn't strictly needed when retraining the FGN or when training from scratch. Figure \ref{fig:decoupled} is an example of an FGN with decoupled bias and center in two dimensional input space. All the experiments performed in later sections were done with decoupled bias and center.

\begin{figure}[!t]
    \centering
    \makebox[\textwidth][c]{
    \begin{subfigure}[t]{0.33\textwidth}
        \includegraphics[width=\textwidth]{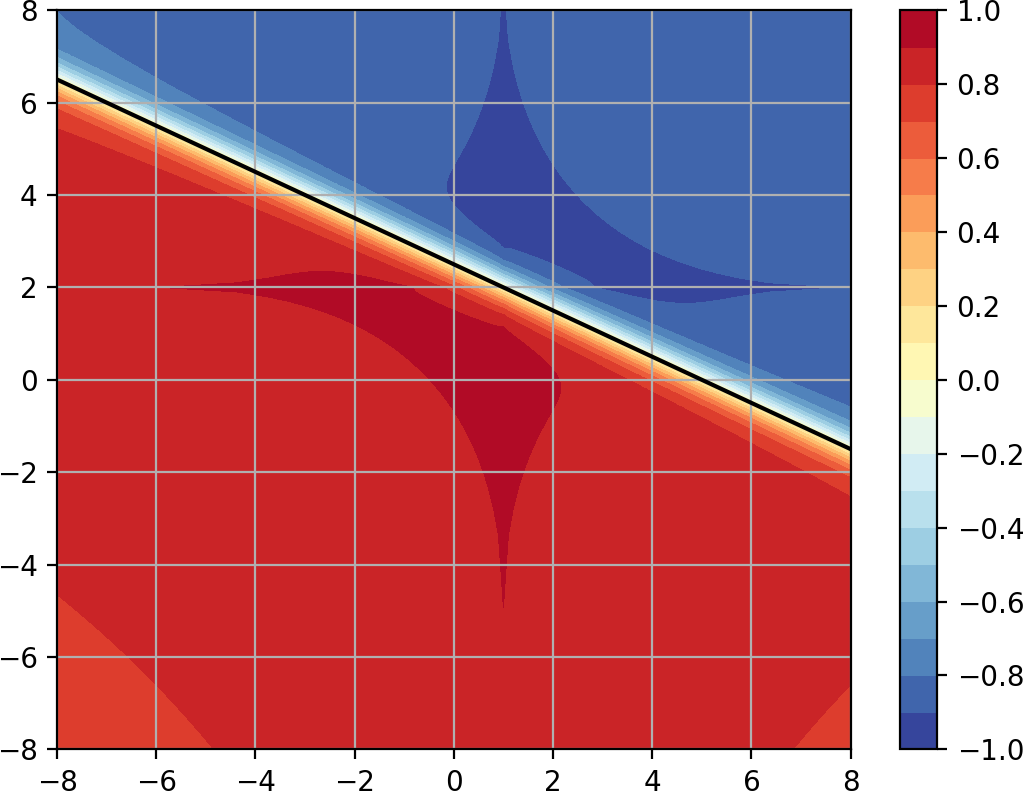}
        \caption*{$p=0.5$}
    \end{subfigure}
    \begin{subfigure}[t]{0.33\textwidth}
        \includegraphics[width=\textwidth]{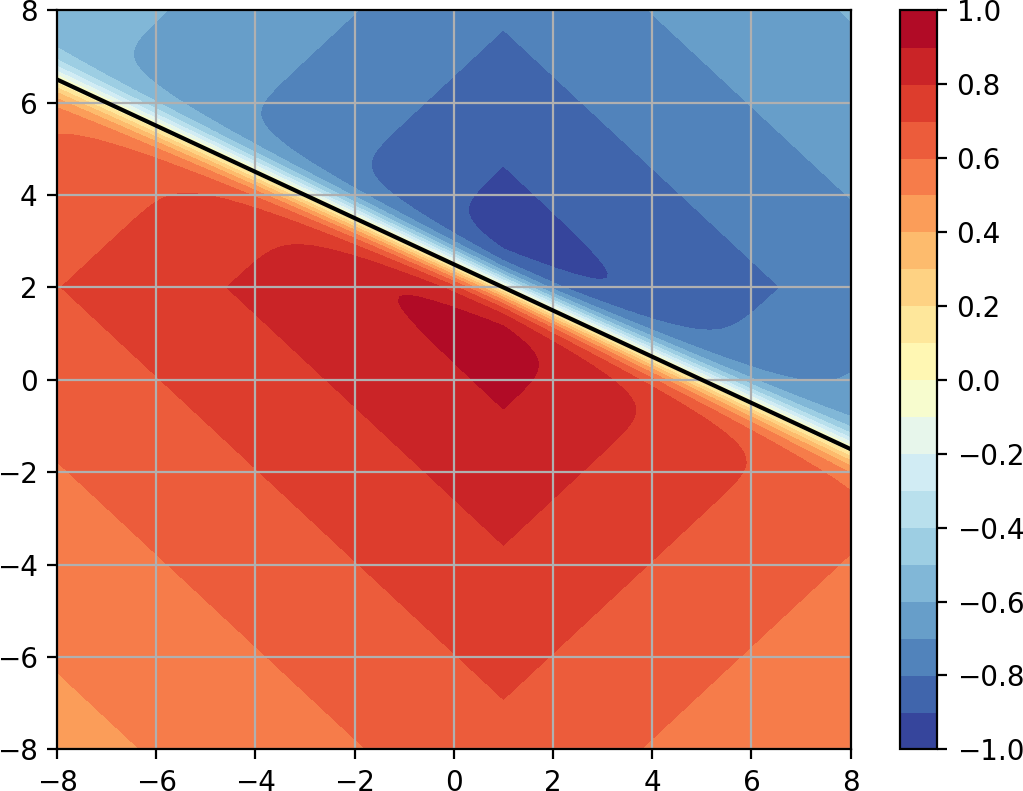}
        \caption*{$p=1.0$}
    \end{subfigure}
    \begin{subfigure}[t]{0.33\textwidth}
        \includegraphics[width=\textwidth]{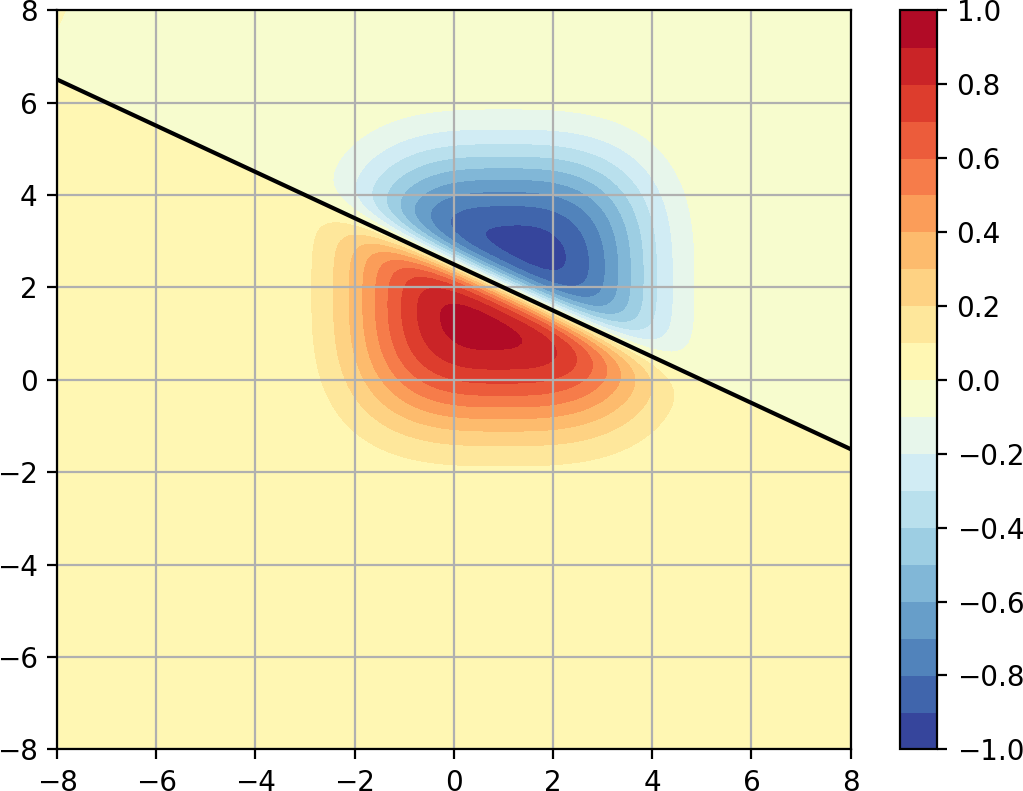}
        \caption*{$p=3.0$}
    \end{subfigure}
    }
    \caption[Various FGN $p$-norm examples]{Examples of how the $p$-norm of the FGN's Gaussian component affects activity.}
    \label{fig:norms}
\end{figure}

\noindent Since distances in higher dimensions behave counter-intuitively, various norms for the Gaussian component $g$ were considered over the Euclidean norm.
\begin{align}
    g = \exp \left( -\frac{1}{\sigma^2}\lVert x_i-c_i \lVert_p \right)
\end{align}
Figure \ref{fig:norms} shows how changing the ordinal $p$ of the norm affects the FGN's output, visualized over a two dimensional input space. Cursory experiments, testing prediction accuracy and resistance to adversarial inputs, have not shown any replicable differences in FGN performance between the various $p$-norms. The Euclidean norm is used in subsequent chapters.

\begin{figure}[!t]
    \centering
    \makebox[\textwidth][c]{
    \begin{subfigure}[t]{0.5\textwidth}
        \includegraphics[width=\textwidth]{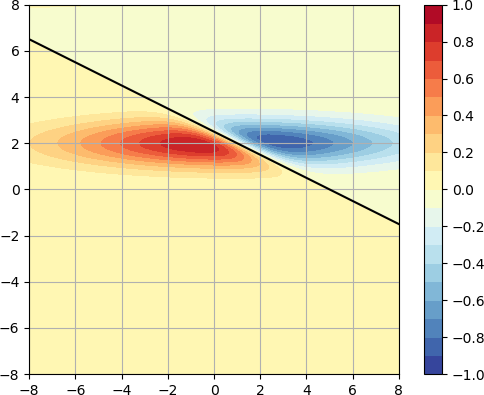}
        \caption*{Diagonal $\Sigma$}
    \end{subfigure}
    \begin{subfigure}[t]{0.5\textwidth}
        \includegraphics[width=\textwidth]{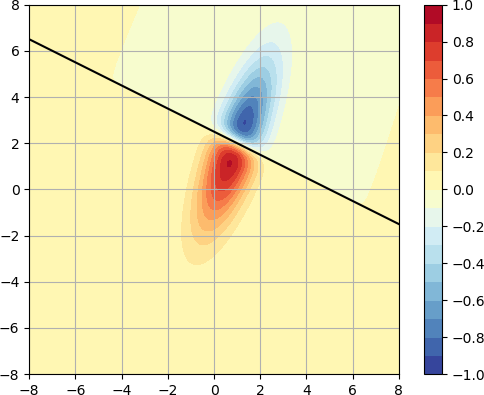}
        \caption*{Full $\Sigma$}
    \end{subfigure}
    }
    \caption[Diagonal and full covariance matrix examples]{Examples of how using a covariance $\Sigma$ matrix, rather than a scalar value $\sigma$, allows the FGN to consider changes along one input dimension to be more important than others.}
    \label{fig:covars}
\end{figure}

\noindent The variance term $\sigma$, which defines how large or small the hyper-sphere of activity is for the Gaussian component $g$, can be modified into a matrix $\Sigma$ to stretch and/or rotate the sphere into an ellipse, which gives the FGN a different variance along each of its inputs.
The Gaussian component $g$ then becomes:
\begin{align}
    g = \exp \left( -(X-C)^T \Sigma^{-1} (X-C) \right)
\end{align}
with $X=[x_i]$ the inputs vector, $C=[c_i]$ the center vector and $\Sigma$ the covariance matrix (positive semi-definite). Figure \ref{fig:covars} gives examples of such $\Sigma$ matrices. Note that a full $\Sigma$ matrix has $N^2$ elements, with $N$ being the number of inputs to the FGN, making the computation of $g$ challenging for larger problem. A diagonal $\Sigma$ matrix is feasible; more memory is required to store the variance vector, but the same number of scalar multiplications is done, except with different scalars instead of the same variance for each FGN input.

\section{Multi-Layer FGN Networks}
\begin{figure}[!t]
    \centering
        \includegraphics[width=\textwidth]{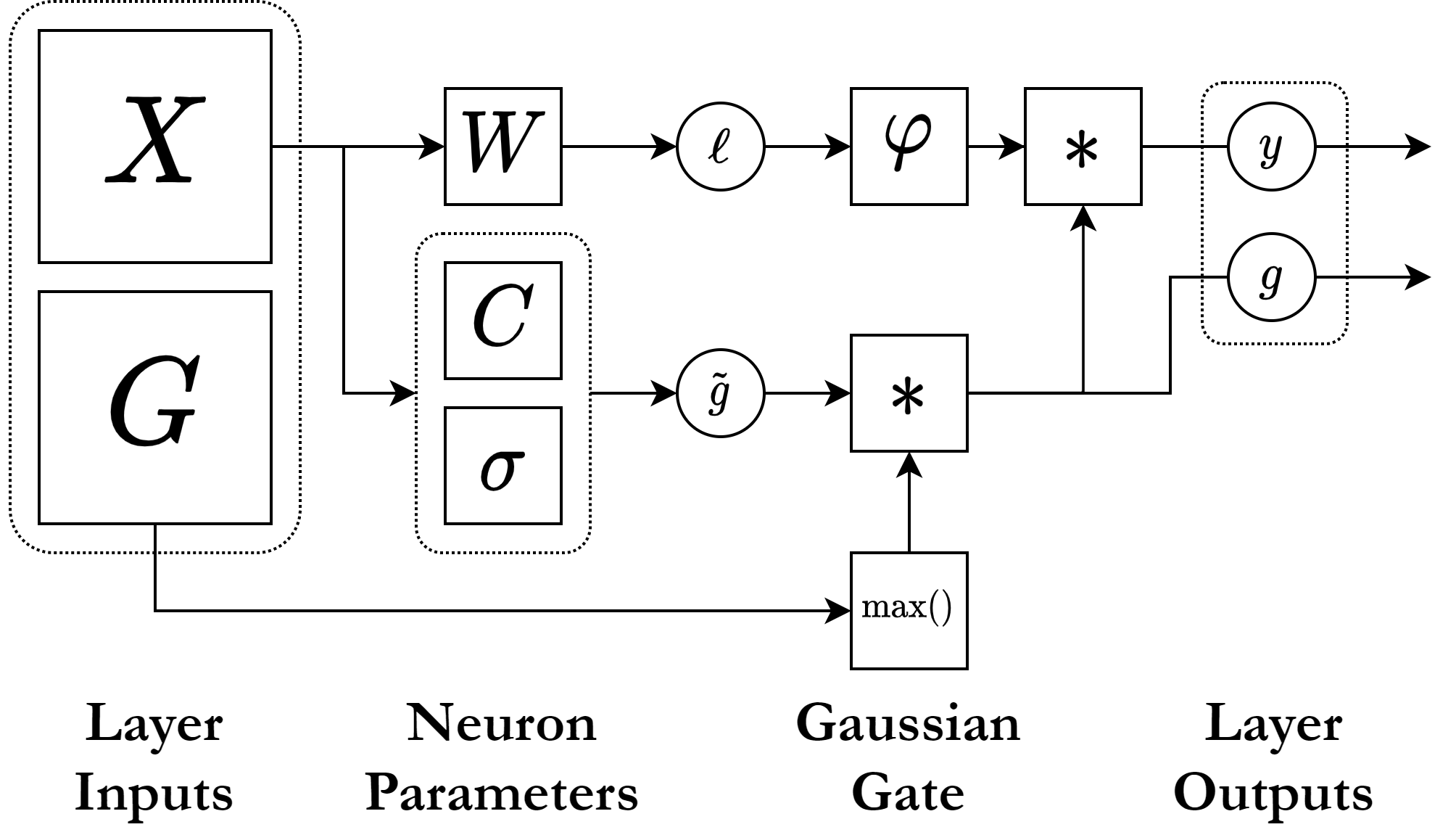}
    \caption[Multi-layer FGN architecture]{Illustration of an FGN's output computation when integrated in a neural network. The added Gaussian gate allows for zero activity due to out-of-range inputs to propagate throughout the network. Both $y$ ang $g$ are outputed to the next layer.}
    \label{fig:fgn-layer}
\end{figure}

\begin{figure}[!t]
    \centering
    \includegraphics[width=0.6\textwidth]{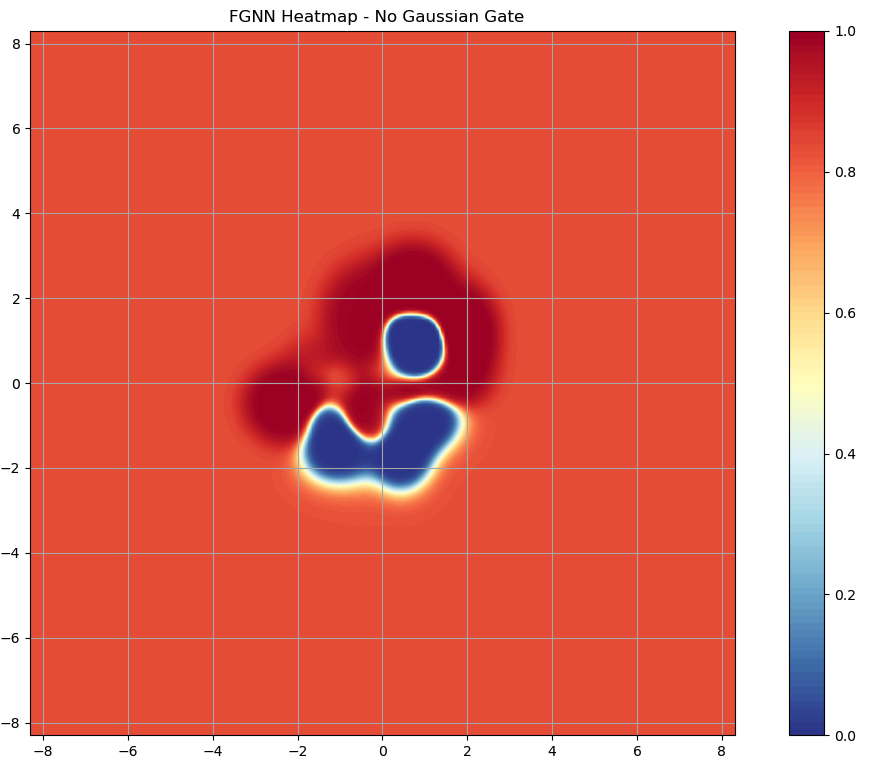}
    \caption[Heatmap without Gaussian Gate]{The activity heatmap of the FGNN when the Gaussian gate is missing. All activity far from the data goes to the same value, but it is not guaranteed to be zero. In this case it has converged to 0.8.}
    \label{fig:toy-2d-no-gate}
\end{figure}

Extending the FGN architecture to neural networks is straightforward save for one detail: the Gaussian component $g$ of an FGN needs to consider the Gaussian components of its input from the previous layer to be able to propagate out-of-range activities. Without this Gaussian gate, only the first layer has zero activity far from the data, and subsequent layer activity far from the data defaults to the FGN bias, not necessarily to zero as desired, as shown in figure \ref{fig:toy-2d-no-gate}.\\
The output $y$ of the $j$-th FGN layer in a FGNN is: 
\begin{align}
    y &=  \varphi(\ell)\cdot g = \varphi \left(\sum_i x_{i} w_{i} \right) \cdot  g\\
    g &= \max(G_{j-1}) \cdot  \exp \left( -\frac{1}{\sigma^2}\sum_{i}(x_i-c_i)^2 \right)
\end{align}
With $x_{i}$ and $G_{j-1}$ the previous layer outputs and Gaussian components. For the initial layer, $\max(G_{j-1})$ should be set to $1$. Figure \ref{fig:fgn-layer} illustrates this process.

\begin{figure}[!t]
    \centering
        \includegraphics[width=0.66\textwidth]{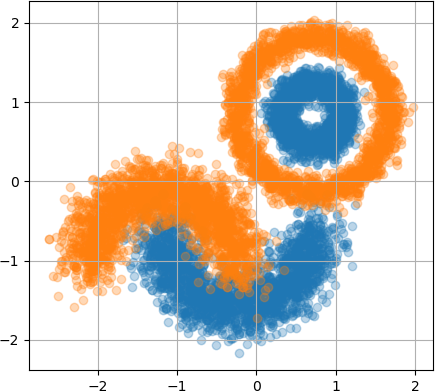}
    \caption[Non-linearly separable toy 2D data]{Two class, two dimensional, non-linearly separable toy dataset}
    \label{fig:toy-2d-data}
\end{figure}

\begin{figure}[!t]
    \centering
    \makebox[\textwidth][c]{
    \begin{subfigure}[t]{0.5\textwidth}
        \includegraphics[width=\textwidth]{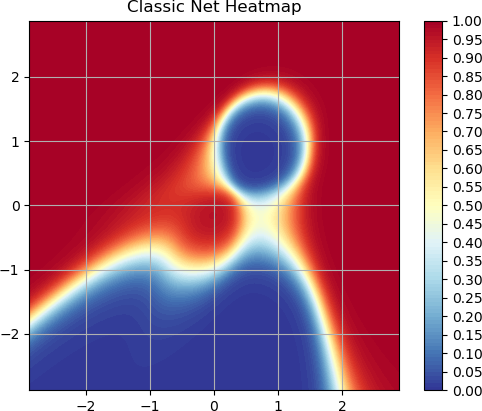}
    \end{subfigure}
    \begin{subfigure}[t]{0.5\textwidth}
        \includegraphics[width=\textwidth]{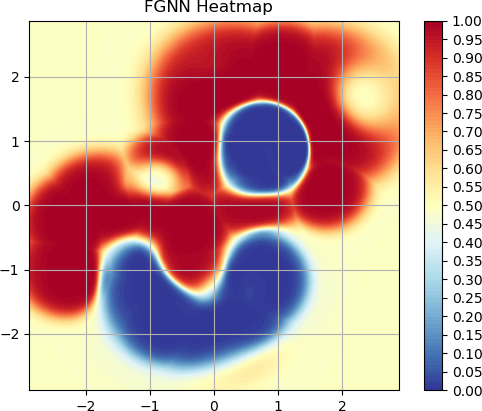}
    \end{subfigure}
    }
    \caption[Activity heatmap comparison: classic vs FGN]{Comparison of the activity heatmap between a classical neural network (left) and the same network using FGNs instead of classical neurons (right). Both networks separate the data shown in figure \ref{fig:toy-2d-data}, but only the classical network makes predictions far from the data.}
    \label{fig:toy-2d-activities}
\end{figure}

\begin{figure}[!t]
    \centering
    \makebox[\textwidth][c]{
    \begin{subfigure}[t]{0.5\textwidth}
        \includegraphics[width=\textwidth]{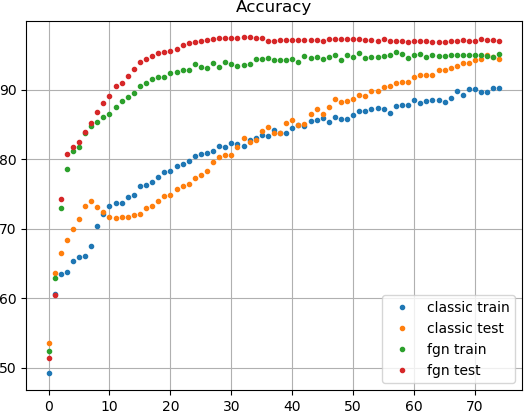}
        \caption*{Classic vs FGN training accuracies}
    \end{subfigure}
    \begin{subfigure}[t]{0.5\textwidth}
        \includegraphics[width=\textwidth]{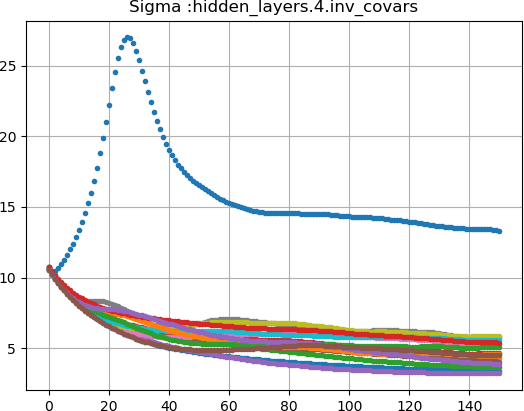}
        \caption*{Changes to the FGNN's variances}
    \end{subfigure}
    }
    \caption[Evolution of accuracy and FGNN variances during training]{For this toy problem, the FGN has better accuracy than the classic model. The plot on the right show how the range of the neurons in the final layer (variance $\sigma^2$) changes during training, most of them shrinking under the added loss pressure shown in \ref{eq:loss}}
    \label{fig:toy-2d-params}
\end{figure}

\newpage
To illustrate the difference in behavior between a neural network built from classical neurons compared to a network built from FGNs, consider the two class, two dimensional, non-linearly separable toy dataset in figure \ref{fig:toy-2d-data}. After training two fully-connected feedforward neural networks, one with FGNs and the other with classical neurons but otherwise identical, figure \ref{fig:toy-2d-activities} shows that, while both networks are able to properly separate the two classes, the FGNN restricts its predictions to zones of the space where the data is present, while the classical network does not. Figure \ref{fig:toy-2d-params} gives additional details on the experiment.

\begin{table}
    \caption{Network parameters}
    \begin{center}
        \begin{tabular}{ll}
        \toprule
        Neurons in hidden layers & 32-16 \\
        Dropout rate & 1/16 \\
        $p$-norm & 2  \\
        Variance & Spherical \\
        First FGN layer center initialization & Random data points\\
        \bottomrule
        \end{tabular}
    \end{center}
    \label{fig:net-params}
\end{table}


\chapter{Experimental Results}
\label{Experimental Results}
\section{FGNNs over MNIST}
\label{FGNNs over MNIST}
\subsection{Experimental Setup}
With the behavior of FGNNs verified on toy data, we now test it on real data, with the goal of providing defense against adversarial attacks. The first dataset tested is MNIST, a collection of handwritten digit images \citet{lecun1998mnist}. Three networks will be compared:
\begin{itemize}
    \item A network trained over MNIST for 50 epochs, built with two fully-connected layers of 64 classical neurons each, with dropout layers with drop rate 0.2 in between each layer, and with a softmax output layer.
    \item A network built from FGNs directly converted from the classic network above, with no retraining, and with a variance $\sigma=10$ large enough to not modify the behavior over the training data.
    \item The same FGNN as above, retrained over the data for a single epoch with a tiny $\lambda=10^{-10}$ loss pressure to shrink the variances $\sigma$.
\end{itemize}
All three networks have over 99\% accuracy over the training data and over 97\% over the validation data. The specific network design choices (size and number of layers) are not relevant to the conclusions drawn, as other designs were tested with similar experimental results.

\subsection{Behavior over Random Images}
\label{mnist over random}
\begin{figure}[!t]
    \centering
    \makebox[\textwidth][c]{
    \begin{subfigure}[t]{0.5\textwidth}
        \includegraphics[width=\textwidth]{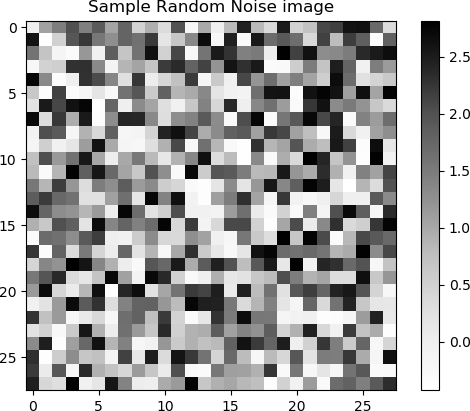}
        \caption*{Fully random pixels}
    \end{subfigure}
    \begin{subfigure}[t]{0.5\textwidth}
        \includegraphics[width=\textwidth]{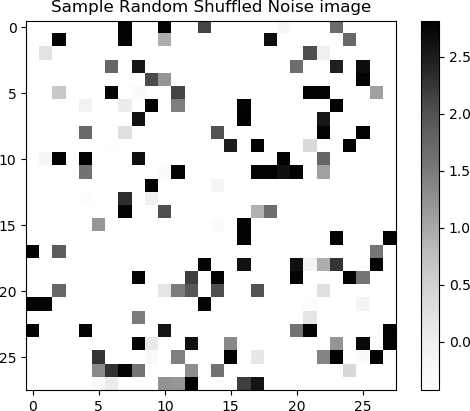}
        \caption*{Shuffled MNIST image}
    \end{subfigure}
    }
    \caption{Samples from the two randomized images datasets.}
    \label{fig:randoms}
\end{figure}

The three networks were tested over two sets of randomized images: one built from images with fully randomized pixel values, which have different mean and variance compared MNIST; and the other built from images consisting of shuffled MNIST images, in which the pixel values are maintained but randomly rearranged, which preserve the mean and variance. Figure \ref{fig:randoms} shows image samples from these two sets. 

\begin{figure}[!t]
    \centering
    \begin{subfigure}[t]{0.6\textwidth}
        \includegraphics[width=\textwidth]{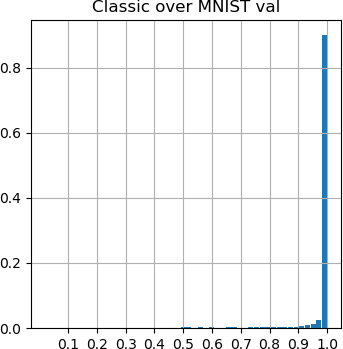}
    \end{subfigure}
    \caption{Histogram of confidences - classic network on MNIST.}
    \label{fig:classic-mnist}
\end{figure}

Not only does the classical network have a high accuracy, it also has a high confidence in its predictions (defined as the maximum value of the softmax function over its output dimensions $Y$, i.e, $\textrm{softmax}(Y_i) = \exp(Y_i) / \sum_k \exp(Y_k) $). Over the validation set, over 99\% of its predictions gave one of the ten digit classes a confidence  greater that $0.5$, i.e., a majority confidence over all the other classes. In fact most of the predictions are well into the [$0.9,1.0$] confidence range, as shown by figure \ref{fig:classic-mnist}.

\newpage

\begin{figure}[!t]
    \centering
    \makebox[\textwidth][c]{
    \begin{subfigure}[t]{0.5\textwidth}
        \includegraphics[width=\textwidth]{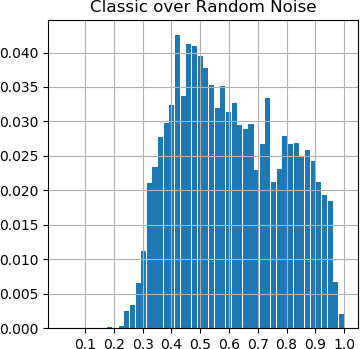}
        \caption*{Over 66\% of confidences $>0.5$}
    \end{subfigure}
    \begin{subfigure}[t]{0.5\textwidth}
        \includegraphics[width=\textwidth]{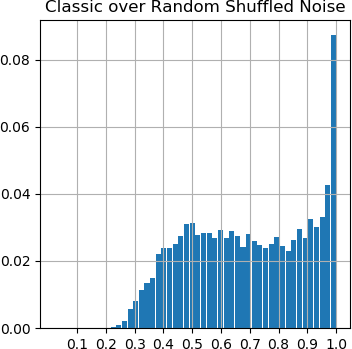}
        \caption*{Over 78\% of confidences $>0.5$}
    \end{subfigure}
    }
    \caption{Histogram of confidences - classic network on random noise}
    \label{fig:classic-hists}
\end{figure}

\begin{figure}[!t]
    \centering
    \makebox[\textwidth][c]{
    \begin{subfigure}[t]{0.5\textwidth}
        \includegraphics[width=\textwidth]{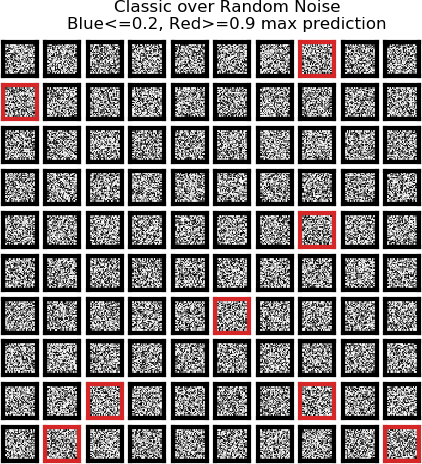}
    \end{subfigure}
    \begin{subfigure}[t]{0.5\textwidth}
        \includegraphics[width=\textwidth]{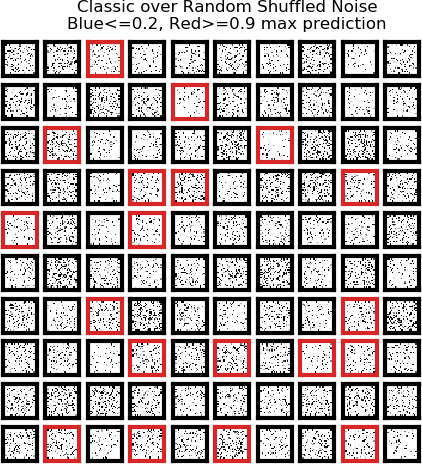}
    \end{subfigure}
    }
    \caption[Sample over-confident predictions made by the classic network]{Sample over-confident predictions made by the classic network. For each red square, one of the model's softmax outputs is above $0.9$ i.e the model is highly confident in its prediction for that image. There are no visible blue squares which would indicate a maximum softmax output below $0.2$ for that image, i.e. the model being close to saying ``I don't know". The values $0.9$ and $0.2$ were selected for illustration purposes and do not hold intrinsic meaning. }
    \label{fig:classic-preds}
\end{figure}

When the same network is tested over the randomized images sets, the histograms of confidences in Figure~\ref{fig:classic-hists} show that a majority of predictions still have above $0.5$ confidence, even though the randomized images are far away in the high dimensional hyper-space from the MNIST images. Figure \ref{fig:classic-preds} shows examples of randomized images, highlighting for illustration purposes those given an MNIST digit class with over 0.9 confidence. This is a replication of results from \citet{szegedy2013intriguing}.

\newpage

\begin{figure}[!t]
    \centering
    \makebox[\textwidth][c]{
    \begin{subfigure}[t]{0.55\textwidth}
        \includegraphics[width=\textwidth]{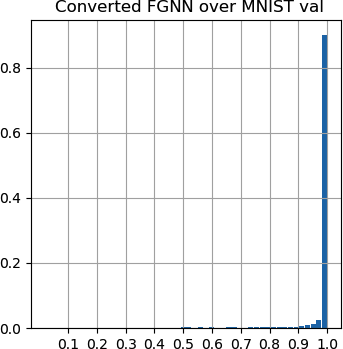}
    \end{subfigure}
    }
    \caption{Histogram of confidences - converted FGNN on MNIST.}
    \label{fig:conv-mnist}
\end{figure}

\begin{figure}[!t]
    \centering
    \makebox[\textwidth][c]{
    \begin{subfigure}[t]{0.5\textwidth}
        \includegraphics[width=\textwidth]{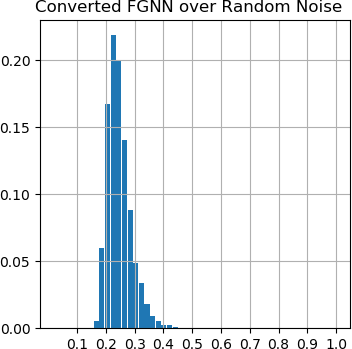}
        \caption*{0.0\% of confidences $>0.5$}
    \end{subfigure}
    \begin{subfigure}[t]{0.5\textwidth}
        \includegraphics[width=\textwidth]{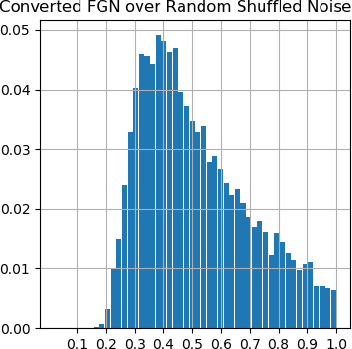}
        \caption*{Over 43\% of confidences $>0.5$}
    \end{subfigure}
    }
    \caption{Histogram of confidences - converted FGNN on random noise}
    \label{fig:converted-hist}
\end{figure}

Corresponding histograms for the converted FGNN are shown in Figure \ref{fig:conv-mnist} and \ref{fig:converted-hist}. By design, the converted FGNN does not modify the behavior of the classic network over the MNIST training data. It outputs identical values and the histogram of its prediction confidences over the validation data looks identical to that of classic network. By contrast, when the converted FGN network is tested over the randomized images sets, the histograms are shifted towards lower values, with fewer confident predictions.
\newpage

\begin{figure}[!t]
    \centering
    \makebox[\textwidth][c]{
    \begin{subfigure}[t]{0.55\textwidth}
        \includegraphics[width=\textwidth]{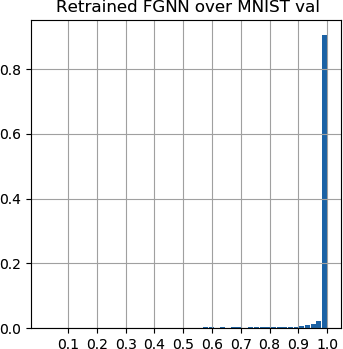}
    \end{subfigure}
    }
    \caption{Histogram of confidences - retrained FGNN network.}
    \label{fig:retrained-mnist}
\end{figure}

\begin{figure}[!h]
    \centering
    \makebox[\textwidth][c]{
    \begin{subfigure}[t]{0.5\textwidth}
        \includegraphics[width=\textwidth]{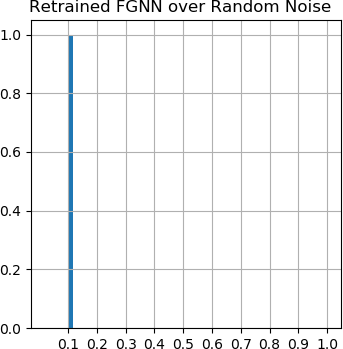}
        \caption*{0.0\% of of confidences $>0.5$}
    \end{subfigure}
    \begin{subfigure}[t]{0.5\textwidth}
        \includegraphics[width=\textwidth]{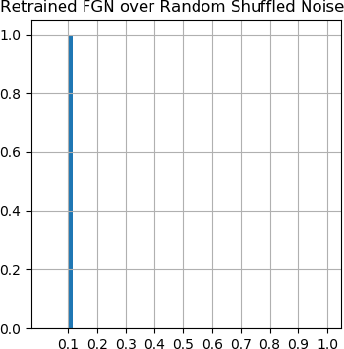}
        \caption*{0.0\% of confidences $>0.5$}
    \end{subfigure}
    }
    \caption{Histogram of confidences - retrained FGNN on MNIST}
    \label{fig:ret-hist}
\end{figure}

Finally, the FGNN is retrained over MNIST for a \emph{single} epoch. The pressure to shrink the variances $\sigma$ of the network's FNGs, added to the loss function, was a $\lambda=10^{-10}$. The overall accuracy remains the same (99\% over training data and 97\% over validation data) and still the network outputs extremely confident predictions over MNIST images. The figure \ref{fig:retrained-mnist} is functionally identical to \ref{fig:classic-mnist}. But when this network is tested over the randomized image sets, almost all the predictions are at $0.1$, as shown in figure \ref{fig:ret-hist}, meaning the retrained FGN network's softmax output is actually all zeros, and making no prediction for these randomized images that are far away in the high dimensional hyper-space from the MNIST images. The retrained FGNN is restricting its activity exclusively to zones of the hyper-space in which the training data are present. These zones are still large enough to generalize from the training data to the validation data. With a value of $\lambda=10^{-10}$, the overall loss increased by under 0.05\%. By comparison the l2-regularization factor of $10^{-5}$ adds $\sim$10\%), indicating that the variance $\sigma$ shrinkage can be a lower priority task than the prediction accuracy and classical l2 weight regularization while still being effective at limiting the range of the FGNN. Retraining for a single epoch compared to the classical networks initial 50 epochs shows that only a fraction of the original works needs to added for an effective FGNN to be trained.

\newpage 

\begin{figure}[!t]
    \centering
    \makebox[\textwidth][c]{
    \begin{subfigure}[t]{0.33\textwidth}
        \includegraphics[width=\textwidth]{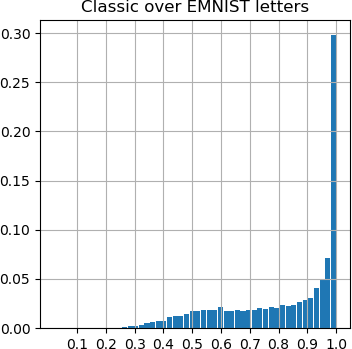}
        \caption*{90\%  of confidences $>0.5$}
    \end{subfigure}
    \begin{subfigure}[t]{0.33\textwidth}
        \includegraphics[width=\textwidth]{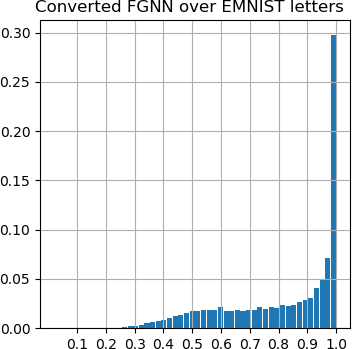}
        \caption*{90\% of confidences $>0.5$}
    \end{subfigure}
    \begin{subfigure}[t]{0.33\textwidth}
        \includegraphics[width=\textwidth]{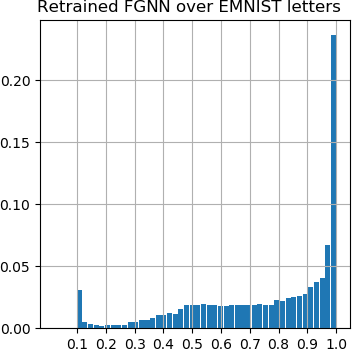}
        \caption*{83\% of confidences $>0.5$}
    \end{subfigure}
    }
    \caption{Histogram of confidences - classic and FGNNs on EMNIST}
    \label{fig:hist-letters}
\end{figure}

\begin{figure}[!t]
    \centering
    \makebox[\textwidth][c]{
        \includegraphics[width=0.33\textwidth]{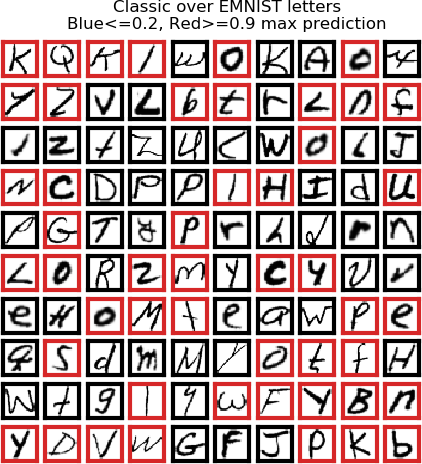}
        \includegraphics[width=0.33\textwidth]{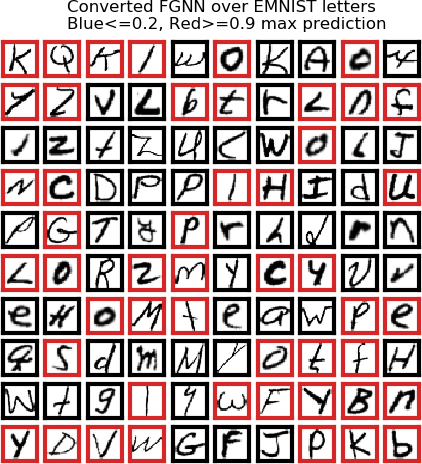}
        \includegraphics[width=0.33\textwidth]{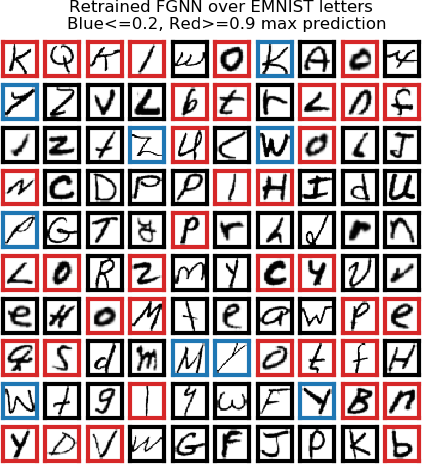}
    }
    \caption[Sample EMNIST predictions - classic and FGNs]{Sample EMNIST images with their prediction confidences given by the different networks. Images of the letter O are identified as a digit 0 with high confidence by all networks, but other images are also assigned a high confidence prediction. Only the retrained FGNN rejects some images as out-of-domain.}
    \label{fig:pred-letters}
\end{figure}

To explore why FGNNs can generalize from training to validation dataset while still rejecting randomly generated images, the three networks are tested on the EMNIST dataset, a collection of handwritten letter images by \citet{cohen2017emnist}. The histograms of the prediction confidences are shown in figure \ref{fig:hist-letters}. These networks were not retrained and do not aim to classify EMNIST images properly.
Unlike over random images, classic and FGNNs behave similarly, with only some images being rejected as out-of-domain by the retrained FGNN. This suggests that FGNs do not simply restrict activity to areas close to training samples, but instead that the hyper-surface of non-zero activity needed to cover the training samples, defined by the smallest variances $\sigma$ possible, also cover both the validation data, and naturally generated yet out-of-domain samples.
Samples images are shown in figure \ref{fig:pred-letters}.

\newpage
\subsection{Protection against FGSM attack}
\label{prot-fgsm}
While protection from out-of-domain predictions is a useful feature of FGNNs, it does not guarantee protection from adversarial attacks. To verify this, the networks were tested against the untargeted FGSM attack first introduced by \citet{goodfellow2014explaining}, which generates adversarial examples by taking a single fixed-size step in the opposite direction of the gradient of the loss with regards to the input image and the true image label (see \ref{eq:fgsm}). To be considered successful an attack must not only change the class of the prediction outputted by the network for this input from correct to incorrect, but also have a confidence in the incorrect  class above 0.5 (i.e., the network has stronger confidence in this class than all others combined). The untargeted version of the attack was chosen for its simplicity.

\begin{figure}[!t]
    \centering
    \makebox[\textwidth][c]{
        \includegraphics[width=1.05\textwidth]{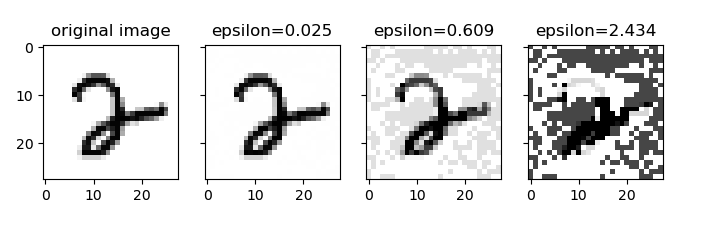}
        }
    \caption[FGSM examples]{An MNIST image and the adversarial images produced by the FGSM attack on the classic network, for various values of step-size $\epsilon$}
    \label{fig:fgsm-exs}
\end{figure}

A parameter of the FGSM attack is $\epsilon$, the size of the fixed-length step, which also corresponds to the maximum amount of distortion allowed to the original image. Larger $\epsilon$ leads to a larger number of successful attacks, but makes the added adversarial noise more noticeable to a human observer. The smallest $\epsilon$ tested is equal to the smallest change possible to an 8-bit pixel. 
In addition to the three networks previously described in Chapter \ref{Experimental Results} (the classical network, the converted FGNN, and the retrained FGNN), a fourth network is tested against FGSM. This network is built from the converted FGNN and is retrained for 100 epochs, the same number of epochs used to train the classical network, and with a much larger pressure $\lambda=10.0$ to shrink the variances $\sigma$ of the FGNs added to the loss. This long-retrained FGNN maintains a 99\%/97\% accuracy during training/validation, with a high confidence in its predictions over MNIST.
The FGSM attack was run over the 10,000 images from the MNIST validation set. Examples of adversarial images are show in figure \ref{fig:fgsm-exs}.
\newpage

\begin{figure}[!t]
    \centering
        \makebox[\textwidth][c]{
        \includegraphics[width=1.1\textwidth]{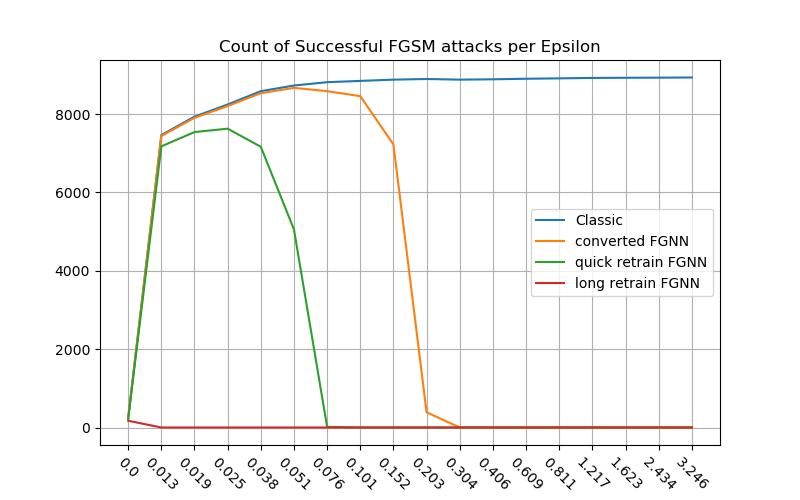}
        }
    \caption[Count of successful FGSM attacks]{Count of successful FGSM attacks (changes the class from correct to incorrect, confidence $>0.5$) among the 10K MNIST validation set images.}
    \label{fig:fgsm-counts}
\end{figure}
\begin{figure}[!t]
    \centering
    \makebox[\textwidth][c]{
        \includegraphics[width=0.5\textwidth]{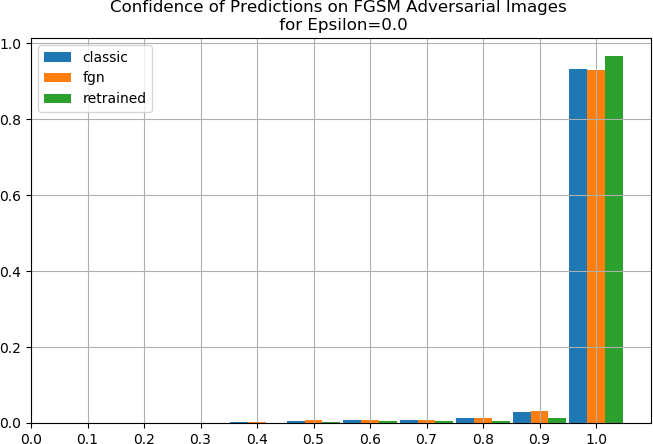}
        \includegraphics[width=0.5\textwidth]{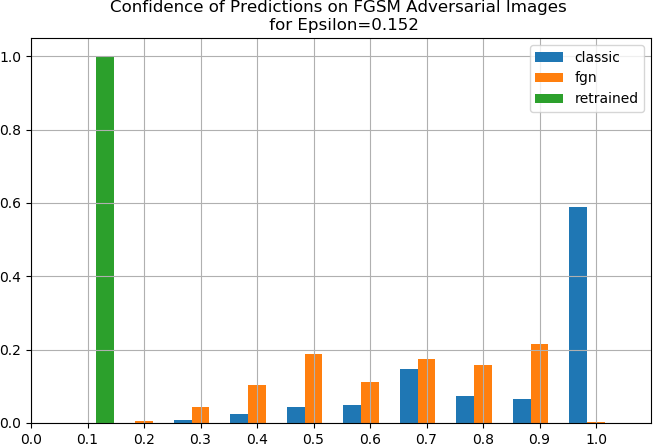}
    }
    \makebox[\textwidth][c]{
        \includegraphics[width=0.5\textwidth]{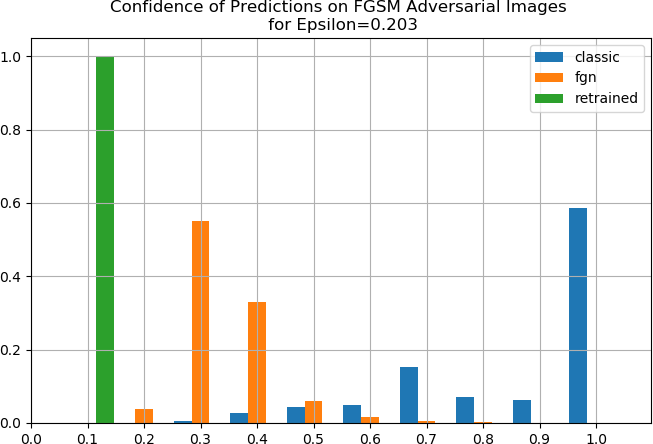}
        \includegraphics[width=0.5\textwidth]{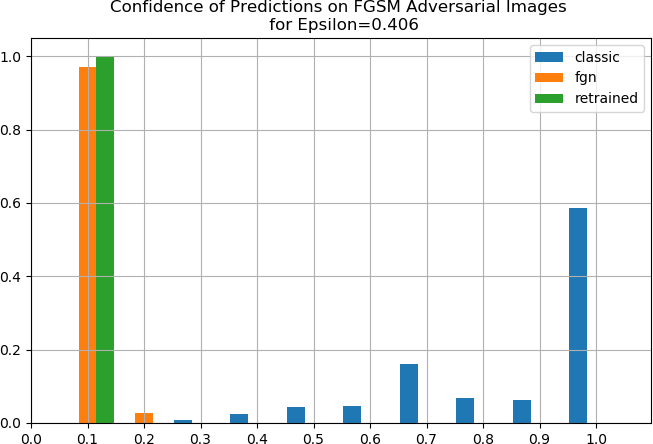}
    }
    \caption[Histograms of confidences on FGSM images]{Histograms of the confidences in the predictions over FGSM adversarial images for the classical network, the converted FGN network, and the long-retrained network. For $\epsilon=0$ the adversarial images are the same as the MNIST images and the networks output predictions with high confidence. For any non-zero $\epsilon$, the long-retrained FGNN outputs all zeros, while the classical network outputs high confidences. As $\epsilon$ increases, the converted FGN network starts behaving differently from the classical network, outputting lower confidences, until it matches the long-retrained FGN network and outputs all zeros. }
    \label{fig:fgsm-compar}
\end{figure}

\noindent Figure \ref{fig:fgsm-counts} show the count of successful FGSM attacks per epsilon, over the 10K MNIST validation set images. The lower the count, the more the network is considered robust to adversarial attacks.
The results  show that FGNNs are more resistant to the FGSM adversarial attack than the classic network. The long-retrained FGNN appears to be impervious to the FGSM attack even for small $\epsilon$, while the classical network remains vulnerable to attacks no matter the $\epsilon$ value. For both the converted and the quick-retrained FGN networks, smaller $\epsilon$ values still allow for a comparable number of successful attacks as the classical network, but past a certain $\epsilon$ value are no longer vulnerable to the FGSM attack. Taking a closer look, all the outputs of the FGNNs on the adversarial images in the cases with no successful attacks are zeros, indicating that these adversarial images are falling outside of the range of these networks' internal FGNs. Figure \ref{fig:fgsm-compar} shows the evolution of the confidence in the adversarial predictions as $\epsilon$ increases.

\newpage 
\subsection{Observations}
\label{Observations}
\begin{figure}[!t]
    \centering
    \makebox[\textwidth][c]{
        \includegraphics[width=0.33\textwidth]{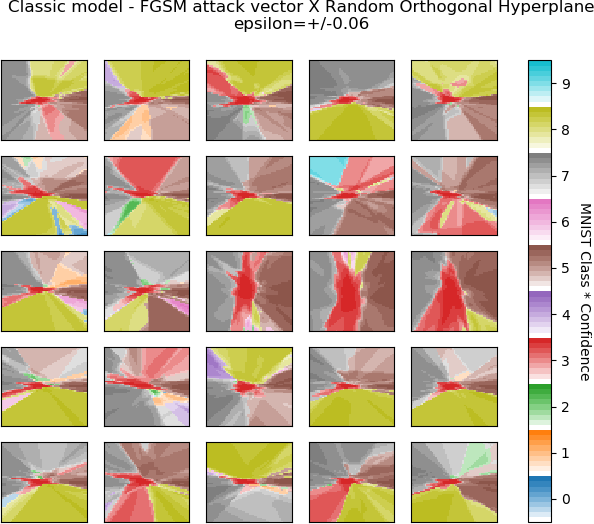}
        \includegraphics[width=0.33\textwidth]{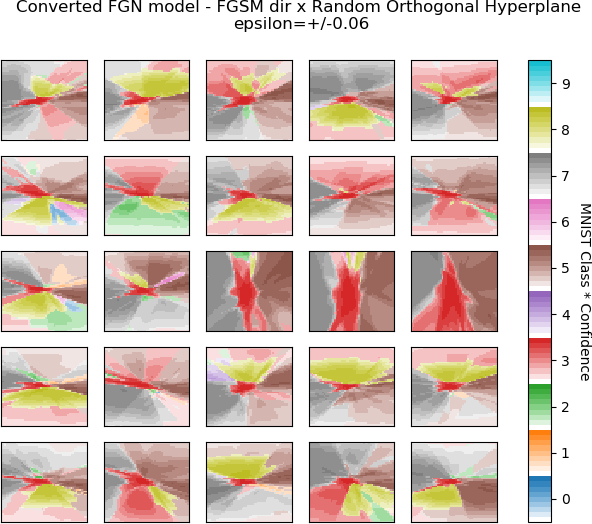}
        \includegraphics[width=0.33\textwidth]{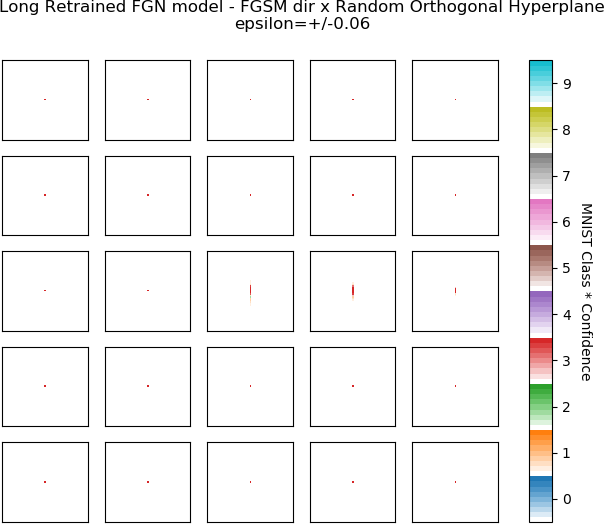}
    }
    \caption[MNIST model boundaries - adversarial cross-section]{Boundaries between predicted classes in the hyper-space around an MNIST example for the classical network, the converted FGNN and the long-retrained FGNN. Each small image is square 2D cross-section of the hyperspace, defined by an MNIST image at the center, the FGSM attack vector on the horizontal axis, and a random orthogonal vector on the vertical axis. The images are $2\epsilon$ wide. Each pixel is colored by the predicted class, with higher color intensity for higher confidence. This is example shows a digit 3 adversarially transformed into a digit 5 for the classical network and converted FGN network.\protect\footnotemark}
    \label{fig:fgsm-bounds}
\end{figure}

These results indicate that FGNNs offer some protection from FGSM attacks. They seem to avoid overfitting, generalizing from training data to validation data, while making no predictions on adversarial images. To explore this seemingly contradictory phenomenon, following \citet{goodfellow2016deep} we visualize the decisions of these networks over various low dimensional manifolds in the high dimensional input space. Figure \ref{fig:fgsm-bounds} plots two dimensional cross-sections of the 784 dimensional MNIST image space, filling the space with colors associated with digit classes as predicted by the networks. The images show the prediction on an MNIST image at the center, moving along the FGSM attack vector to the right, and a randomly chosen vector orthogonal to the attack vector up. The half-width of the cross section was chosen to be $\epsilon=0.06$ as that is when the classical network and converted FGN network start behaving differently. 

For the classical network, boundaries between the original predicted class and every other class are present, indicating the existence of many adversarial examples within the $\epsilon$-hypersphere around the image. The converted FGNN exhibits the same boundaries but with weaker confidence as we move away from the original image, making adversarial examples rarer. And the long-retrained FGNN makes no predictions anywhere except at the original image, making adversarial examples non-existent.

\begin{figure}[!t]
    \centering
    \makebox[\textwidth][c]{
        \includegraphics[width=0.33\textwidth]{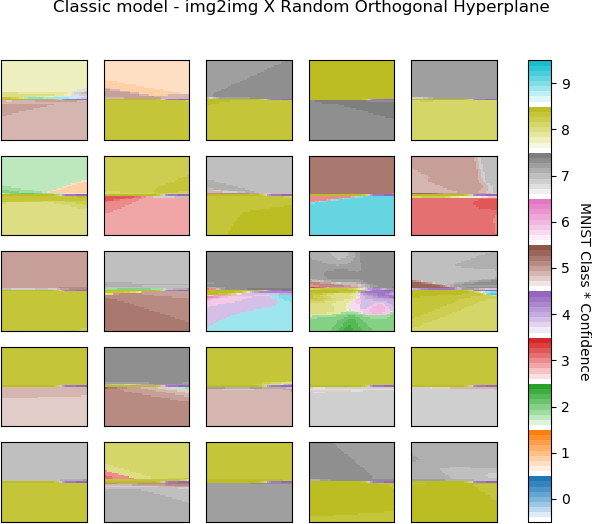}
        \includegraphics[width=0.33\textwidth]{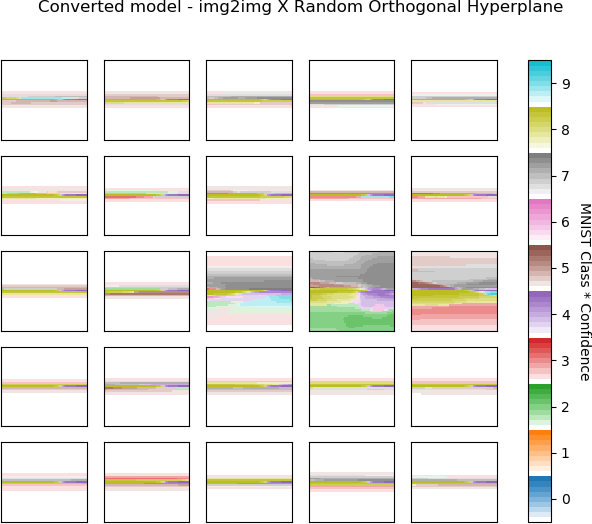}
        \includegraphics[width=0.33\textwidth]{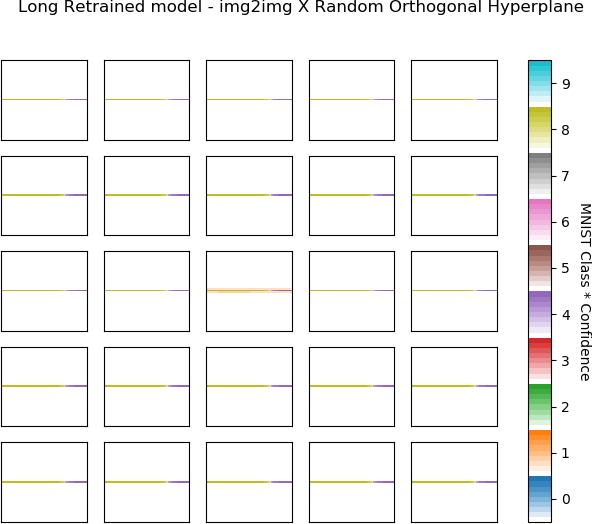}
    }
    \caption[MNIST model boundaries - sample-to-sample cross-section]{Boundaries between predicted classes in the hyper-space between to MNIST images for the classical network, the converted FGNN, and the long-retrained FGNN. Each small image is a square 2D cross-section of the hyperspace, defined by a vector from one MNIST image at the leftmost center pixel to another MNIST image a the rightmost center pixel, and a random orthogonal vector on the vertical axis. Each pixel is colored by the predicted class, with higher color intensity for higher confidence. This example shows a digit 8 linearly transformed into a digit 4.}
    \label{fig:fgsm-img2img}
\end{figure}

The lack of predictions around the MNIST image by the long-retrained FGNN seems to imply that the model is overfitting to the training data, but the high accuracy on the validation data already contradicts this. Figure \ref{fig:fgsm-img2img}, which plots cross-sections of the hyperspace while moving from one MNIST image to another, shows that both FGNNs make predictions continuously along the image-to-image vector, and rarely nearby, while the classical network makes confident predictions over the entire nearby space. It's notable that there are no areas of zero-valued predictions in between the images.\footnotetext{Larger version of images are available in the addendum.} 

\newpage
\section{Comparison to Bayesian Neural Networks}
\label{Bayesian}
Because FGNNs were designed to not output any predictions over samples far from the training data, it makes sense to compare them another technique with a similar property: Bayesian Neural Networks (BNNs). 
BNNs attach a probability distribution to the model parameters, in our case the weights and biases of the underlying neural net, and use samples weight values to make predictions. To produce a prediction, multiple forward passes are computed, each with a new sampled set of weights and biases. Instead of a single set of output values (in our case, softmaxes associated with one of the ten MNIST digit classes), multiple sets are computed which represent a probability distribution over output values. From these distribution we can estimate confidence and uncertainty in each output class. The properties lead to BNNs being able to say \emph{``I don't know."}. A review of BNN theory and practical training methods can be found in \citep{wong2020fast} and \citep{jospin2022hands}.
We use the Pyro toolbox \citep{bingham2019pyro}\footnote{Available at \url{https://github.com/pyro-ppl/pyro}} to implement BNNs.

\subsection{BNNs over random images}
\label{bnn-noise}
\begin{figure}[!t]
    \centering
    \makebox[\textwidth][c]{
    \includegraphics[width=0.13\textwidth]{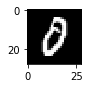}
    \includegraphics[width=0.87\textwidth]{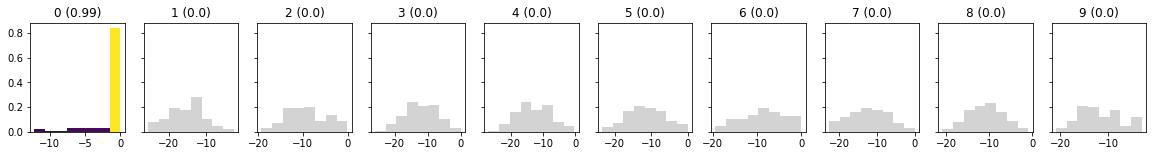}
    }
    \caption[BNN histogram for each class]{The histograms of the log-softmax of the BNN outputs at each digit class, over the 100 weight samplings. The number above each histogram is the median probability given by the BNN for that digit. The histogram is colored yellow if this median probability is above 0.5}
    \label{fig:bnn-ex-dist}
\end{figure}

To evaluate the behavior of a BNN over random noise, and to look at its resistance against FGSM adversarial images, the architecture of the classical model trained in Section~\ref{FGNNs over MNIST} (2 fully-connected layers of 64 classical neurons each) were used as the base for a BNN trained until it reached similar accuracy on the validation set (approximately 97\%). Retraining the BNN to reach this accuracy took significantly more epochs, 5000 on average vs 100 for the classical and FGN networks.
Figure \ref{fig:bnn-ex-dist} shows the histograms of the outputs of the BNN after sampling from the weight distributions 100 times, for a given MNIST image. 

\begin{figure}[!t]
    \centering
    \makebox[\textwidth][c]{
    \includegraphics[width=0.13\textwidth]{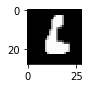}
    \includegraphics[width=0.87\textwidth]{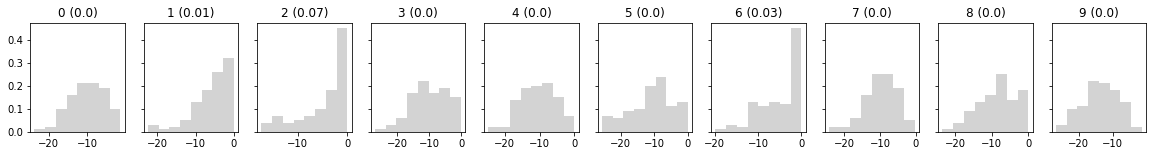}
    }
    \makebox[\textwidth][c]{
    \includegraphics[width=0.13\textwidth]{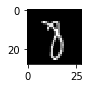}
    \includegraphics[width=0.87\textwidth]{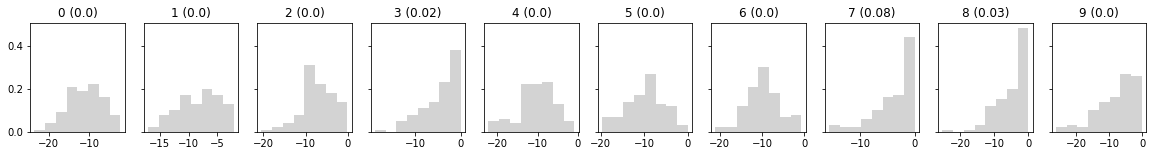}
    }
    \makebox[\textwidth][c]{
    \includegraphics[width=0.13\textwidth]{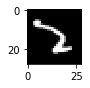}
    \includegraphics[width=0.87\textwidth]{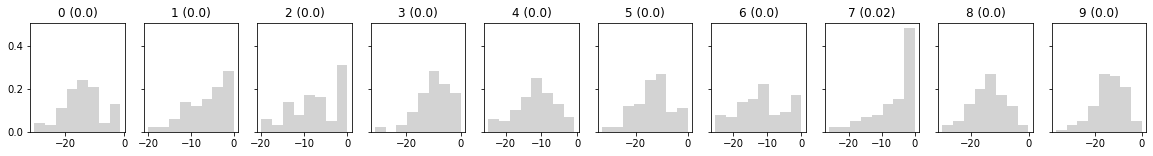}
    }
    \caption[Images rejected by the BNN]{Examples of images rejected by the BNN and their associated histograms.}
    \label{fig:bnn-rejected}
\end{figure}

\newpage
When the BNN is allowed to reject images in which none of the median values of the histograms surpass 0.2, the accuracy over the non-rejected images goes up to 98\% over the accepted images, and approximately 2\% of images are rejected (about 200 out of the 10000 in the validation set). Figure \ref{fig:bnn-rejected} shows examples of rejected images.

\begin{figure}[!t]
    \centering
    \begin{subfigure}[t]{0.66\textwidth}
        \includegraphics[width=\textwidth]{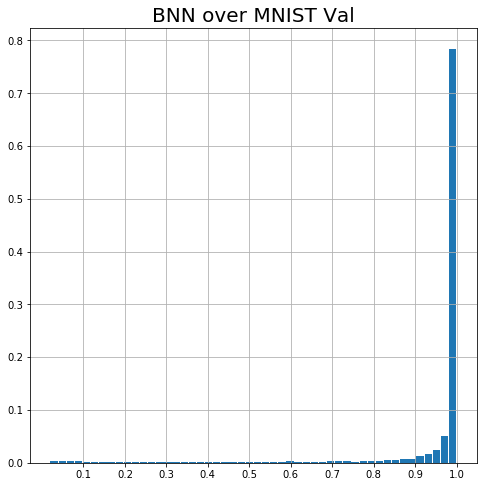}
    \end{subfigure}
    \caption[Histogram of confidences - BNN on MNIST]{Histogram of confidences over MNIST validation set - Bayesian Neural Network.}
    \label{fig:bnn-hist-mnist}
\end{figure}

\begin{figure}[!t]
    \centering
    \makebox[\textwidth][c]{
    \begin{subfigure}[t]{0.5\textwidth}
        \includegraphics[width=\textwidth]{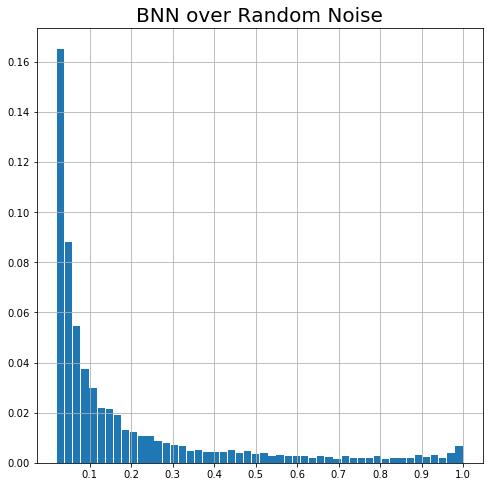}
        \caption*{7\% of confidences $>0.5$}
    \end{subfigure}
    \begin{subfigure}[t]{0.5\textwidth}
        \includegraphics[width=\textwidth]{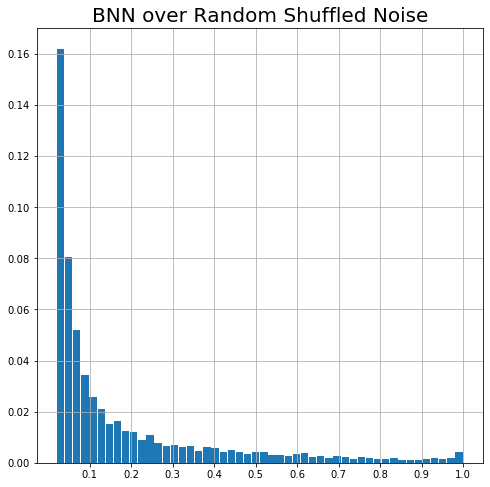}
        \caption*{6\% of confidences $>0.5$}
    \end{subfigure}
    }
    \caption[Histogram of confidences - BNN on random noise]{Histogram of confidences over random noise - Bayesian Neural Network.}
    \label{fig:bnn-random-hists}
\end{figure}

 Just like the classical neural network, the BNN is highly confident in its predictions: over 95\% of its predictions gave one of the ten digit classes a confidence greater that 0.5, and the histogram of these predictions, shown in figure \ref{fig:bnn-hist-mnist} looks indistinguishable from the classical neural network's histogram shown in figure \ref{fig:classic-mnist}
When evaluated over the fully random and shuffled random datasets described in Section \ref{FGNNs over MNIST}, the BNN behaves much better than the classical neural network. On both datasets only 6-7\% of random images produce a confidence greater that 0.5. The histograms are shown in Figure \ref{fig:bnn-random-hists}.
In contrast to FGNNs, where a model re-trained for a single epoch was able to reject every random image as out-of-domain (c.f., Figure~\ref{fig:ret-hist}), the BNN still accepts a small but measurable number of random images. 

\subsection{BNN over adversarial images from FGSM}
\label{bnn-fgsm}
\begin{figure}[!t]
    \centering
        \includegraphics[width=\textwidth]{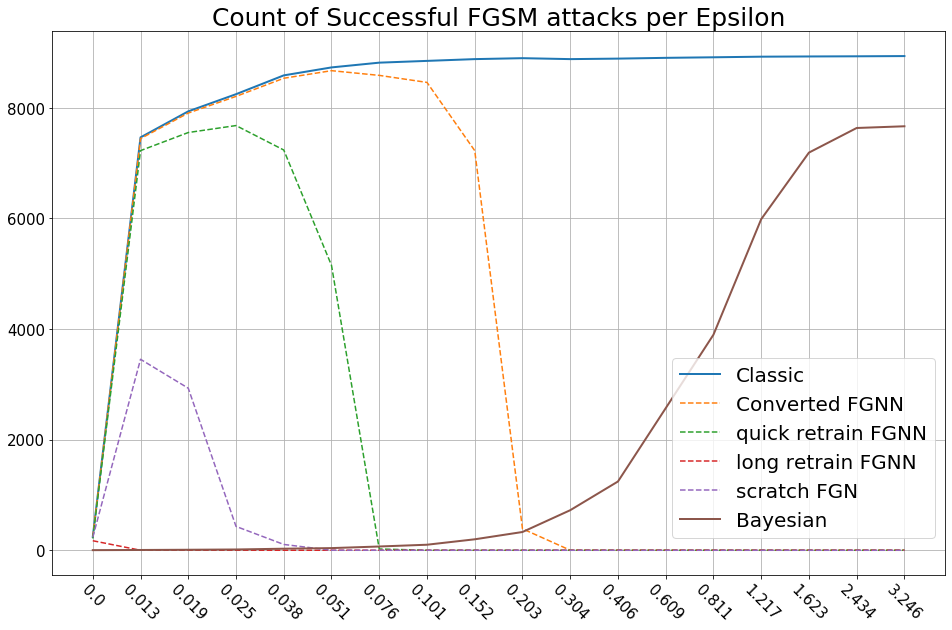}
    \caption[Comparison of the count of successful FGSM attacks - BNN]{Count of successful FGSM attacks (changes the class, confidence~$>0.5$) among the 10K MNIST validation set images for the Bayesian neural network, and the classic and FGN networks.}
    \label{fig:bnn-counts}
\end{figure}

Finally, the adversarial images produced by the FGSM attack on the BNN's corresponding classic neural network (the same that were produced for figure \ref{prot-fgsm}), over the 10K MNIST validation set images, were transferred and tested on the BNN. Successful attacks per epsilon are shown in figure \ref{fig:bnn-counts}.

 The BNN proved extremely resilient, not allowing any successful attacks (changes the class, confidence $>0.5$) until the allowed distortion was over $1/32$ of the maximum pixel range ($\epsilon>0.152$). This behavior contrasts with that of FGNNs. The FGNNs that were simply converted, or retrained for a single epoch, were vulnerable to attacks with small epsilon, but eventually rejected every adversarial images as out-of-domain for larger distortions.
 
\begin{figure}[!t]
    \centering
    \makebox[\textwidth][c]{
        \includegraphics[width=0.5\textwidth]{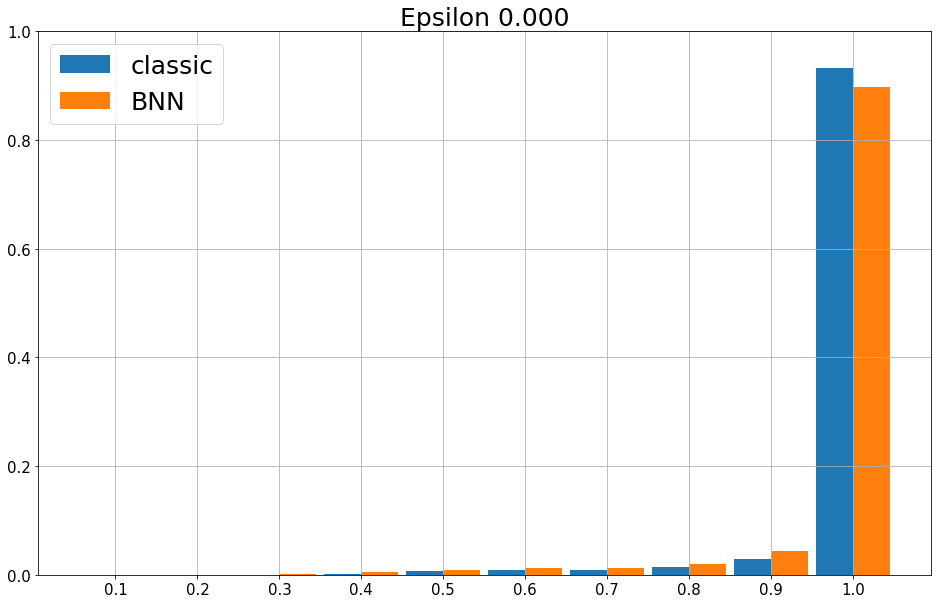}
        \includegraphics[width=0.5\textwidth]{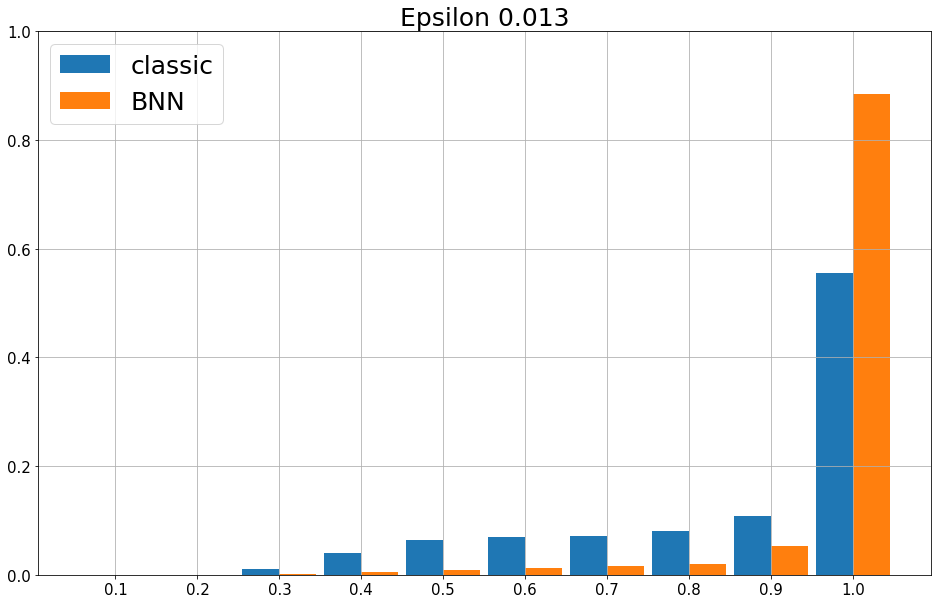}
    }
    \makebox[\textwidth][c]{
         \includegraphics[width=0.5\textwidth]{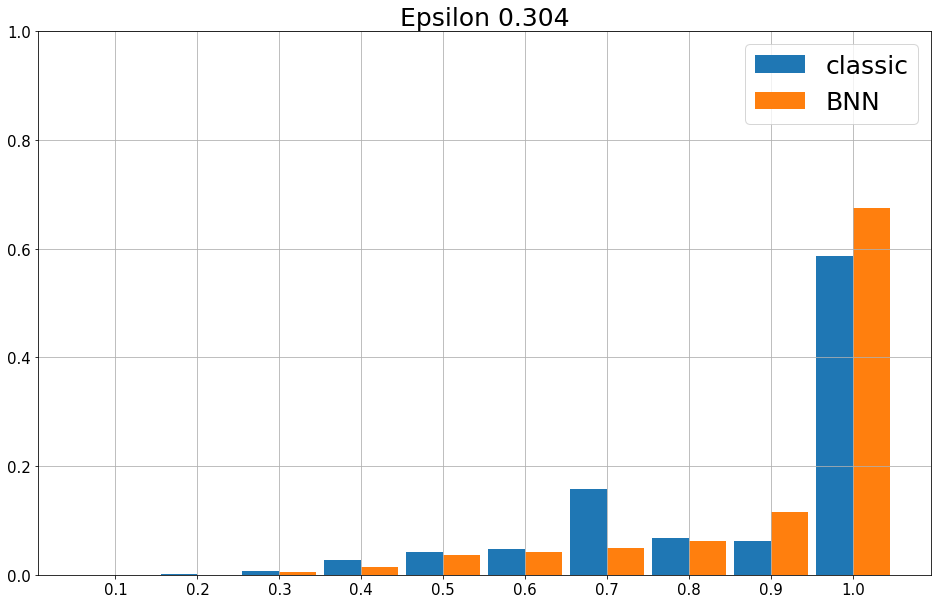}
        \includegraphics[width=0.5\textwidth]{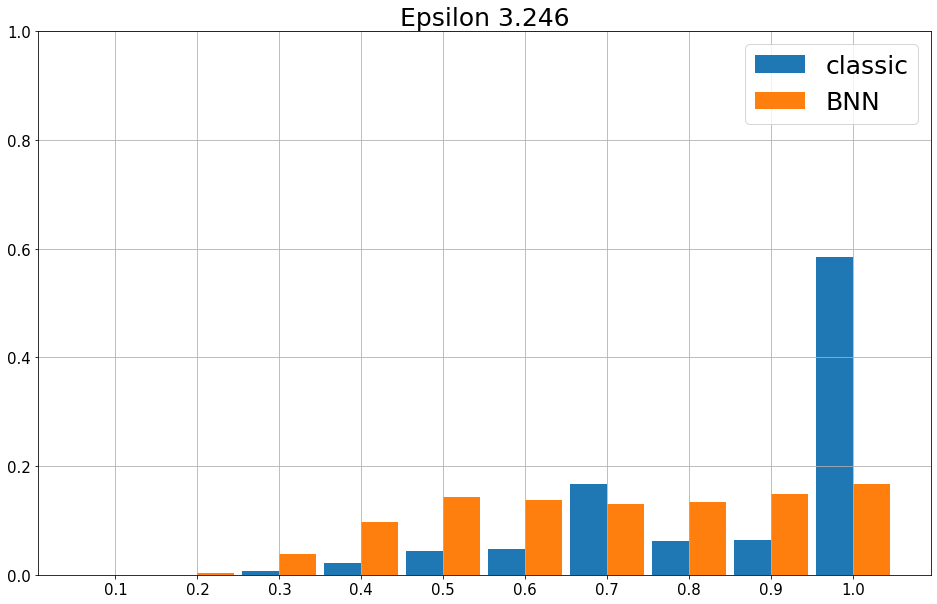}
    }
    \caption[Histogram confidences - Classic vs BNN on FGSM]{Histograms of the confidences in the predictions over FGSM adversarial images for the classical network and the converted and retrained BNN network. While the BNN eventually rejects more adversarial images than the classical neural network, it never matches the FGNNs and outputs all zeros as shown in figure \ref{fig:fgsm-compar}.}
    \label{fig:bnn-fgsm-compar}
\end{figure}

 The difference in behavior over adversarial images of BNNs compared to FGNNs can also be seen in the histograms of their prediction confidences (the maximum softmax of their outputs) over the varying epsilons, shown in figure \ref{fig:bnn-fgsm-compar}. While the BNN eventually rejects more adversarial images than the classical neural network, it never matches the FGNNs and outputs all zeros as shown in figure \ref{fig:fgsm-compar}.
It's notable that the BNN's histogram for maximally distorted adversarial images ($\epsilon=3.246$) still shows many highly confident predictions, even though the BNN's histogram over random images shows few such high confidences (fig \ref{bnn-noise}). This validates the finding of \citet{goodfellow2014explaining} that adversarial images are not random noise.

Combining the above results, we conclude that FGNNs and BNNs defend against FGSM adversarial images through different mechanisms. FGNNs do not distinguish between only slightly distorted adversarial images and normal images unless retrained for a long time, in which case the image is rejected entirely and not recognized as any class. However they will always reject images that are distorted past a threshold. Meanwhile BNNs exhibit the opposite behavior. They can correctly recognize adversarial images that are only slightly distorted, but cannot reject highly distorted images.
It should also be noted that FGNNs and BNNs differ in how they add additional computational complexity to the networks. FGNs require the computation of the Gaussian component of each neuron, and optionally extra training epochs; while BNNs require extra training, and multiple samplings per forward pass.

\section{Convolutional FGNs over SPEECHCOMMANDS}
The results above indicate that FGNNs provide certain neural networks with resistance to certain adversarial attacks, but to further validate their usefulness, FGNNs still need to be tested on on datasets other than MNIST, with a network architecture other that feedforward, against attacks other than FGSM.

\subsection{Description of the Data and Models}
 The SPEECHCOMMANDS dataset by \citet{DBLP:journals/corr/abs-1804-03209} consists of over 100\,000 one-second long utterances of 30 short words from different speakers. The recordings are provided as wav files sampled at 16 kHz. The words were chosen to be a mix of potential commands (`up', `down'), numbers, and common words that could confuse the model (`tree', `dog'). This dataset was chosen to contrast audio against MNIST's visual data, and for the relative simplicity of its audio task. Isolated keyword detection directly relates to the output layer of the neural network, whereas other tasks might require upstream and downstream audio processing unrelated to the testing of the neural network against adversarial attacks. For example continuous text-to-speech models might change word predictions based on prior probabilities estimated by a language model, which fall outside the scope of analyzing an attack on a specific neural network.
 Though in the last decade, large attention-based models \citet{bahdanau2016end} have demonstrated that neural networks are capable of end-to-end automatic speech recognition, highlighting the importance of understanding how such large networks respond to adversarial attacks, without needed to attack each different subsystem separately.

The neural networks trained on the SPEECHCOMMANDS tasks were modeled after the M5 architecture by \citet{dai2017very}. This architecture was chosen because of its reported performance on the task (71\% accuracy, state of the art at the time, though we achieve over 80\%), the presence of convolutional layers within the network, and the PyTorch library's providing open-source code for the training of such networks. Convolutional layers have proven themselves to be one of the most useful neural network layer variants, especially since Alex Krizhevsky et. al. \citet{krizhevsky2012imagenet} used them to win the 2012 ImageNet challenge by over 10 percentage points. Current state of the art on this task reaches around 97-98\%, as reported by \citet{vygon2021learning}.

A single one dimensional convolution layer is essentially one neuron slid across the input vector according to the kernel size, stride and dilation parameters. As such, a convolutional layer of classical neurons can be converted to FGN neurons in the same way described in section \ref{eq:fgn}.  For a convolutional FGN layer with neuron weights $W=[w_0, w_1 \ldots w_k]$, centers $C=[c_0, c_1, \ldots c_k]$ and variance $\sigma$; with convolutional parameters stride $s$, kernel size $k$ and dilation $d$; the $i$-th value of output $Y$ for an input vector $X$ of length $n$,  is:
\begin{align}
    Y_i &= \varphi(\ell_i) \cdot g_i \\
    \ell_i &= \sum_{k=0}^{K} w_k \cdot Z_{ik} \\
    g_i &= \exp \left(-\frac{1}{\sigma^2}\sum_{k=0}^{K}(Z_{ik}-C_k)^2 \right) \\
    Z_{ik} &= X_{(i+k)(s+d)} 
\end{align}
The size of the output Y is $\lfloor 1 + (n-d(k-1) -1)/s \rfloor$.

\begin{figure}[!h]
    \centering
    \makebox[\textwidth][c]{
    \begin{subfigure}[t]{0.5\textwidth}
        \includegraphics[width=\textwidth]{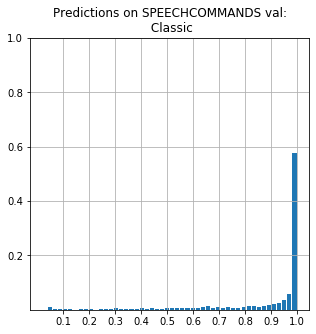}
        \caption*{Over 91\% of confidences $>0.5$}
    \end{subfigure}
    \begin{subfigure}[t]{0.5\textwidth}
        \includegraphics[width=\textwidth]{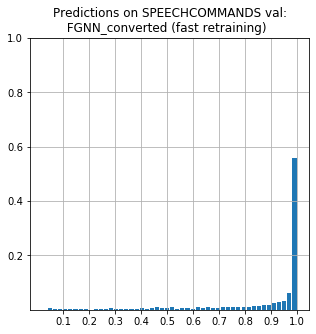}
        \caption*{Over 89\% of confidences $>0.5$}
    \end{subfigure}
    }
    \makebox[\textwidth][c]{
    \begin{subfigure}[t]{0.5\textwidth}
        \includegraphics[width=\textwidth]{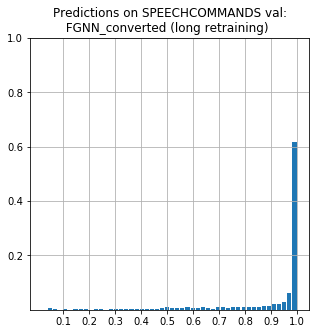}
        \caption*{Over 90\% of confidences $>0.5$}
    \end{subfigure}
    \begin{subfigure}[t]{0.5\textwidth}
        \includegraphics[width=\textwidth]{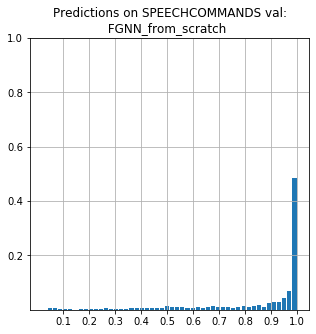}
        \caption*{Over 93\% of confidences $>0.5$}
    \end{subfigure}
    }
    \caption[Histogram of confidences - SPEECHCOMMANDS]{Histogram of confidences of the networks on the SPEECHCOMMANDS validation dataset}
    \label{fig:hists-speechcommands}
\end{figure}

Similarly to the experiment on MNIST, a total of five networks were created: the original M5 network with classic neurons trained for 21 epochs, an FGNN with weights simply converted from the original network without any additional training, the same converted FGNN trained for 1 epoch, the same converted FGNN trained for 21 epochs, and an FGNN trained from scratch for 21 epochs. Though the goal is not to obtain state of the art performance on the task, it is important that all networks perform similarly to be able to compare their resistance to adversarial attacks. All networks fall within the 87-92\% and 84-86\% accuracy ranges on the training and testing sets respectively, and all networks show high confidence in their predictions (shown in figure \ref{fig:hists-speechcommands}). Note that because the converted FGNN does not change the behavior of the classic network on SPEECHCOMMANDS data, their histograms are identical, and thus only one is shown.
\newpage
\newpage

\subsection{Behavior over White Noise}
\begin{figure}[!h]
    \centering
    \makebox[\textwidth][c]{
    \begin{subfigure}[t]{0.5\textwidth}
        \includegraphics[width=\textwidth]{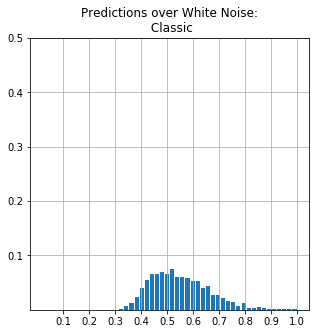}
        \caption*{Over 63\% of confidences $>0.5$}
    \end{subfigure}
    \begin{subfigure}[t]{0.5\textwidth}
        \includegraphics[width=\textwidth]{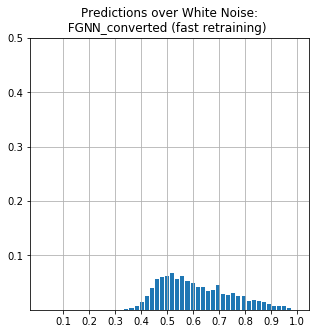}
        \caption*{Over 76\% of confidences $>0.5$}
    \end{subfigure}
    }
    \makebox[\textwidth][c]{
    \begin{subfigure}[t]{0.5\textwidth}
        \includegraphics[width=\textwidth]{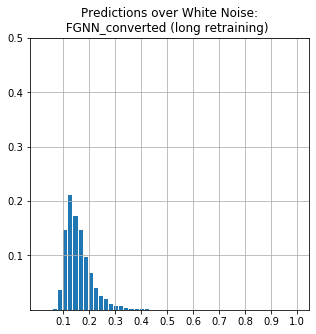}
        \caption*{Under 1\% of confidences $>0.5$}
    \end{subfigure}
    \begin{subfigure}[t]{0.5\textwidth}
        \includegraphics[width=\textwidth]{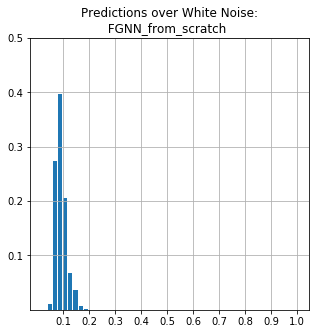}
        \caption*{0\% of confidences $>0.5$}
    \end{subfigure}
    }
    \caption[Histogram of confidences - white noise]{Histogram of confidences of the networks over white noise}
    \label{fig:hists-white-noise}
\end{figure}

In an experiment similar to that described in section \ref{mnist over random}, the 5 networks were tested on white noise. The results are shown in figure \ref{fig:hists-white-noise} (The histogram of confidences of the converted FGNN with no retraining is identical to that of the classic network, and is not shown). As expected based on the results over MNIST, the classic network still makes a substantial amount of confident predictions even though the inputs are completely random. Meanwhile the two FGNN that have been retrained for 21 epochs show essentially no confident predictions. However unlike the MNIST results, the FGNN retrained for a single epoch does make confident predictions, indicating that the variances $\sigma$ define an area of the input space with non-zero activity still large enough to cover all the generated white noise, and that method of initializing the FGNN covariance parameters during conversion could be improved. 
The numerical values of the white noise generated for this experiment fell within the minimum and maximum bounds of the sound samples of the training dataset. In a quick follow-up experiment, these minimum and maximum numerical bounds were expanded by $10^6$, meaning that the networks were given inputs with some values never seen during training. In this case the classic network made 100\% of predictions with confidence over 0.5, while every FGNN, including the one converted FGNN with no retraining and the converted FGNN retrained for 1 epoch, only made null predictions. This result shows that the FGNN does have an limited area of the input space with non-zero activity, while the classic network does not. 

\newpage

\subsection{Protection against Attacks}
The two attacks chosen to test the networks resistance were the Carlini-Wagner attack \citep{carlini2017evaluating} and the projected gradient descent (PGD) attack \citep{madry2019deep}. PGD can be easily understood as an iterative version of the FGSM attack, taking multiple steps along the adversarial gradient, while staying within the constraint boundaries after each step (see equation \ref{eq:pgd}).
Based on their experimental results, investigating the landscape of local maxima for multiple starting points of the attacks, \citet{madry2019deep} argue that PGD is in a sense the ``ultimate" first-order adversarial attack, which is why it was chosen to test FGNs.

The Carlini-Wagner (CW) attack was designed to minimize a loss function that was found to empirically produce strong adversarial examples, picked from several candidates Carlini and Wagner found promising. Given input $x$ with model output vector $y$ as logits, pick a target class $\hat{t}$ (different than the real class for $x$) and search using gradient descent for the adversarial $w$ that minimizes:
\begin{align}
    & w = \left \lVert \frac{1}{2} \left( \tanh(w)+1 \right) - x \right \rVert_2^2+c \cdot f \left( \frac{1}{2}(\tanh(w)+1) \right) \\
    & \textrm{with } f(x) = \max \left( \max(y_i : i \neq \hat{t}) - y_{\hat{t}} \right) \\ 
    & \textrm{such that } \lVert w \rVert_2 < \epsilon
\end{align}
The constant $c$ controls attack success probability vs distance to $x$ trade-off. As in the original paper \citep{carlini2017evaluating}, binary search was used to find the optimal value of $c$ per attack attempt.

In our experiment, we applied the CW and PGD attacks with varying epsilon parameters. In the CW attack, the epsilon value defines the maximum distortion allowed measured by $L_2$ distance with the original sample, whereas in the PGD attack, epsilon defines the step size ($1/30$ of epsilon) taken during each iteration of PGD, for a maximum of 50 steps.

\begin{figure}[!t]
    \centering
        \includegraphics[width=\textwidth]{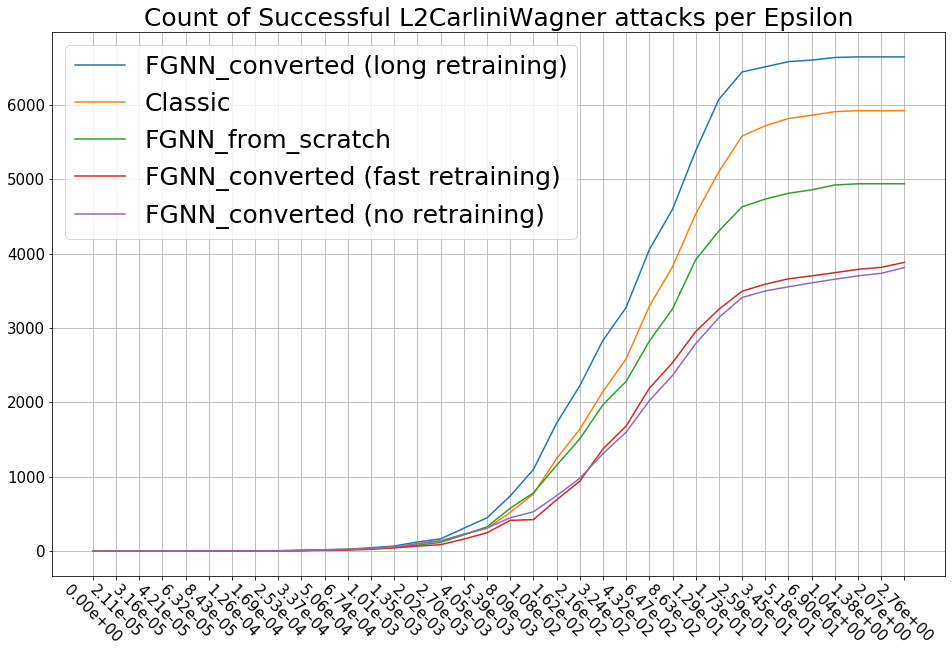}
    \caption[Count of successful Carlini-Wagner attacks]{Count of successful Carlini-Wagner attacks (changes the class, confidence~$>0.5$) per epsilon among the 10K SPEECHCOMMANDS validation audio samples.}
    \label{fig:cw-counts}
\end{figure}

Figure \ref{fig:cw-counts} shows the number of successful Carlini-Wagner attacks per epsilon. Notably, all FGNN perform better than the original classic model, except for the converted FGNN that was re-trained for 21 epochs. This perplexing result is difficult to explain and warrants further exploration. Unlike the FGSM attacks on MNIST, FGNN do not eventually reject every sample, for realistic epsilon values.

\newpage

\begin{figure}[!t]
    \centering
        \includegraphics[width=\textwidth]{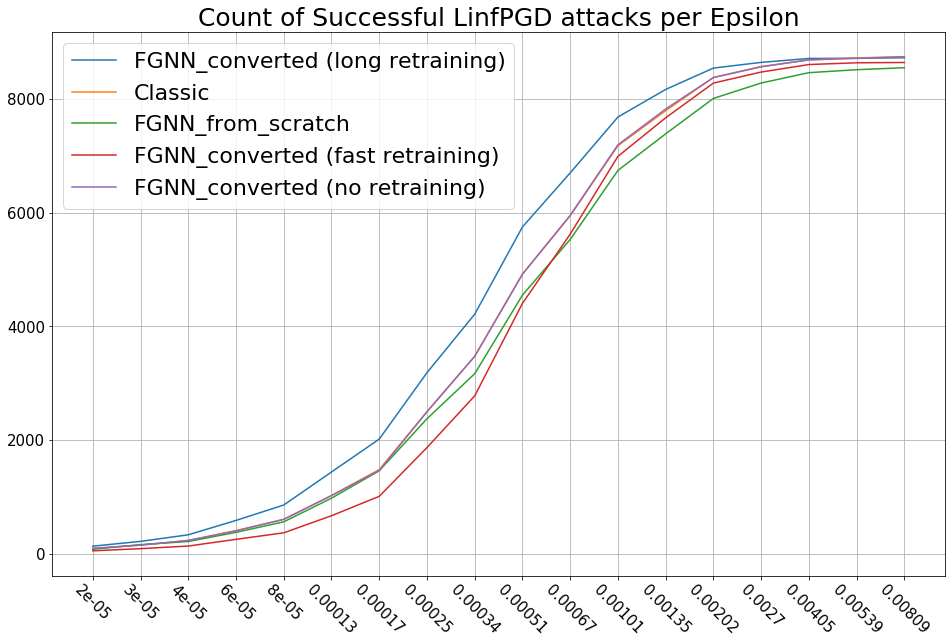}
    \caption[Count of successful PGD attacks]{Count of successful Projected Gradient Descent attacks (changes the class, confidence~$>0.5$) per epsilon among the 10K SPEECHCOMMANDS validation audio samples.}
    \label{fig:pgd-counts}
\end{figure}

Figure \ref{fig:pgd-counts} shows the results of the PGD attack. Finally, unlike in previous experiments, the adversarial attack was strong enough to fool the FGN networks as reliably as the classic model.

\begin{figure}[!t]
    \centering
        \includegraphics[width=\textwidth]{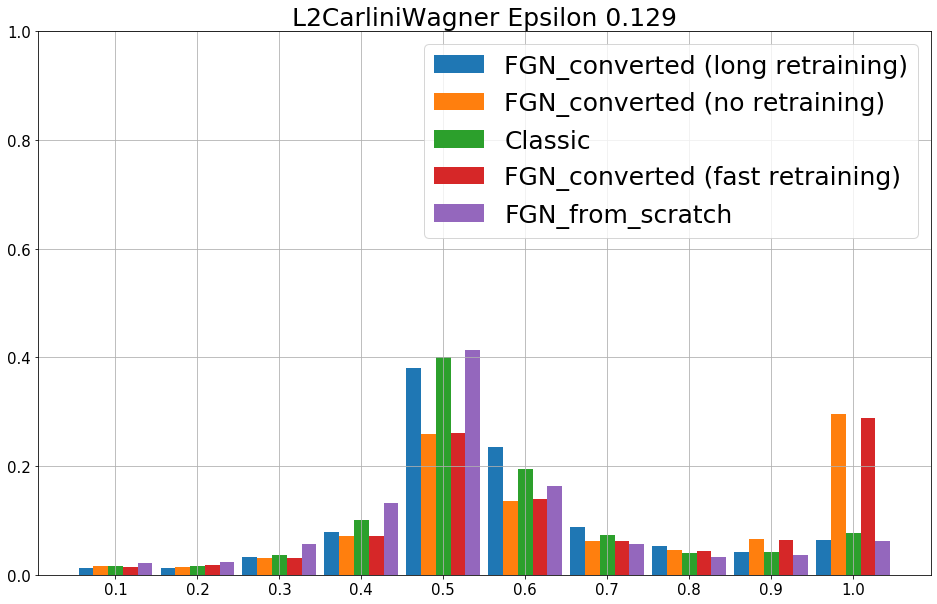}
        \includegraphics[width=\textwidth]{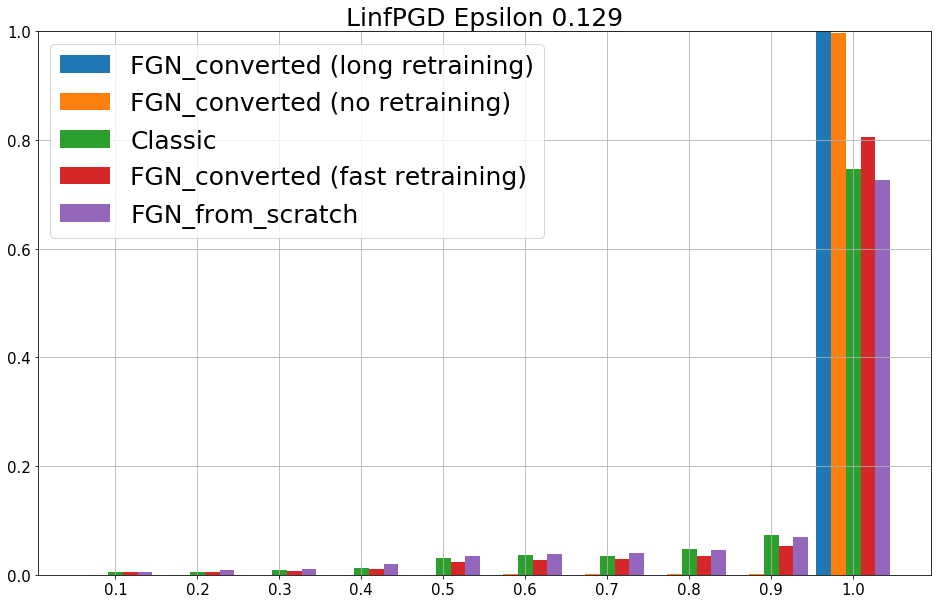}
    \caption[Confidence Histograms for CW and PGD attacks]{}
    \label{fig:adv-confs}
\end{figure}

\newpage
To further explore the differences and similarity of behaviors, figure \ref{fig:adv-confs} shows the histograms of confidences of the classic and FGN networks on the adversarial samples for both the Carlini-Wagner and the PGD attacks. 

For the Carlini-Wagner attack, the two FGNNs with the smallest amount of retraining have the highest confidence in their adversarial predictions, but those are the two networks with the lowest successful adversarial count as seen in \ref{fig:cw-counts}, indicating that the adversarial examples generated by the attack fail to change the class of the original sample. By contrast, FGNNs that have been trained or retrained for longer behave more similarly to the classic network.

In contrast, for the PGD attack, the two FGNNs that behave differently than the classic network are the converted models with no and long retraining. The highest possible confidence in the successful adversarial samples shown by those two networks could indicates that the FGN architecture actually helps the PGD attack, however this is not corroborated by the behavior of the FGNN converted and trained for a single epoch or the FGNN trained from scratch.

Further statistical analysis of the differences between adversarial examples created from classic and FGN networks, such as distribution of distances between original and associated adversarial example, or separating the histograms between the different types of unsuccessful attacks (failed to change the class vs failed to produce high confidence), could help shed further light on the differences in behavior between the models and the failure to defend against the PGD attack.
Testing across multiple model training runs is also needed to verify that the behaviors are not model specific.

\chapter{Conclusion and Future Work}


In this work, the novel Finite Gaussian Neuron architecture is introduced by adapting Radial Basis Function networks to multiple-layers that can be trained using back-propagation.

We find that FGNs are an effective tool for building artificial neural networks that do not make false-positive predictions over random-valued inputs.
Furthermore, networks built using FGNs are shown to protect against the FGSM adversarial attack, comparing favorably to the existing defense of Bayesian neural networks, and offer some resistance against the Carlini-Wagner attack.

A unique property of FGNNs, when compared to other adversarial defenses, is that an existing feed-forward or convolutional neural network can  be converted to the FGN architecture without requiring re-training of the network while maintaining identical behavior. Our finding also shows that this conversion increases the resistance of the network to adversarial attacks.

By design, re-training FGNNs increases their resistance to random inputs. While retraining also increases their resistance to the FGSM attack, it seems to be detrimental to defending against the more complex Carlini-Wagner attack. This counter-intuitive behavior should be explored in future work.

The behavior of FGNNs in the context of high-dimensional vector spaces also warrants further investigation. How are properly trained FGNs in section \ref{FGNNs over MNIST} able to generalize to unseen real images correctly while being able to avoid making predictions over random images? Why are the class boundaries shown in section \ref{Observations} such that FGNNs make predictions in between MNIST image but not around them?
FGNs put into question the reliance of neural networks on the artificial neuron's piece-wise linearity, and invite more research in trainable non-linear neuronal architectures.

We also show that FGNs do not provide defense against the stronger adversarial PGD attack. Exploring how FGNs behave in other settings (black-box attacks, adversarial attack transferability) also warrants further investigation. The strength of the PGD attack suggests that changes artificial neuron's architecture will not be enough to protect from adversarial attacks.  

Finally FGNs highlight the flexibility of back-propagation algorithm in training neural network variants with specific requirements.

\backmatter

\chapter{Appendix}
\begin{figure}
    \centering
    \makebox[\textwidth][c]{
        \includegraphics[width=1.0\textwidth]{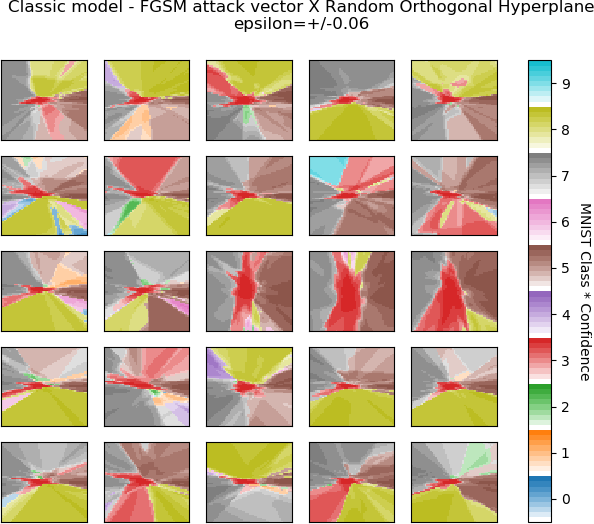}}
    \caption*{Larger version of images in figure \ref{fig:fgsm-bounds}. Boundaries between predicted classes in the hyper-space around an MNIST example for the classical network.}
\end{figure}
\begin{figure}
    \centering
     \makebox[\textwidth][c]{
     \includegraphics[width=1.0\textwidth]{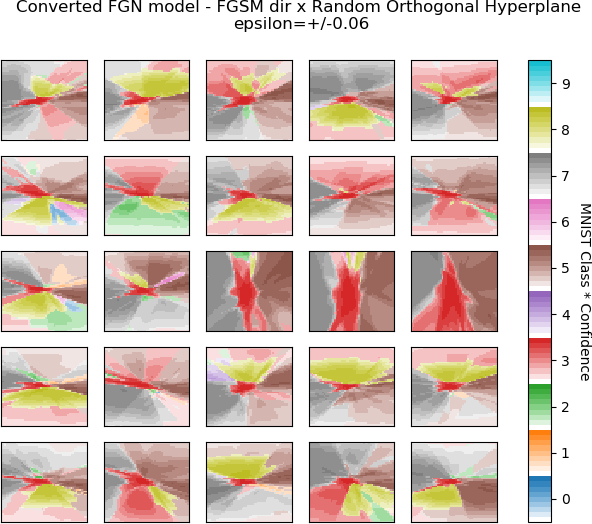}}
    \caption*{Larger version of images in figure \ref{fig:fgsm-bounds}. Boundaries between predicted classes in the hyper-space around an MNIST example for the converted FGNN network.}
\end{figure}
\begin{figure}
    \centering
     \makebox[\textwidth][c]{
     \includegraphics[width=1.0\textwidth]{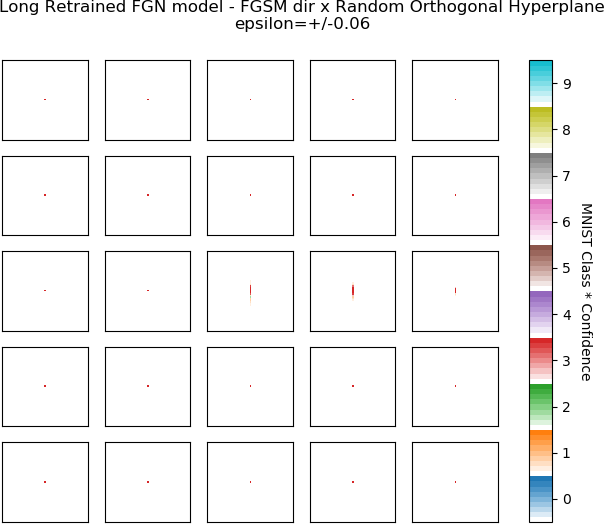}}
    \caption*{Larger version of images in figure \ref{fig:fgsm-bounds}.Boundaries between predicted classes in the hyper-space around an MNIST example for the long-retrained FGNN network.}
\end{figure}
\begin{figure}
    \centering
     \makebox[\textwidth][c]{
     \includegraphics[width=1.0\textwidth]{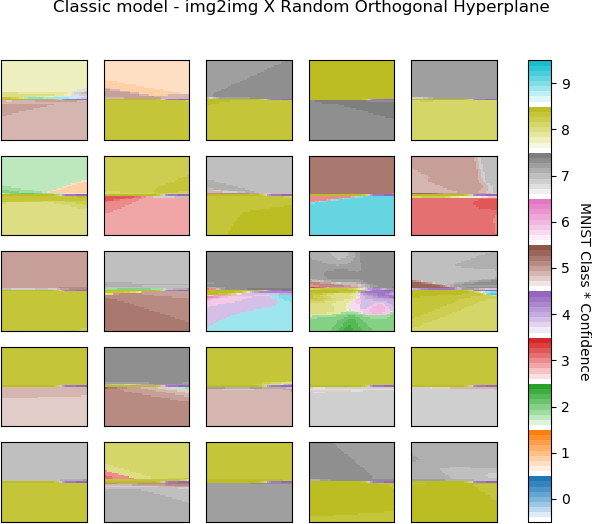}}
    \caption*{Larger version of images in figure \ref{fig:fgsm-img2img}. Boundaries between predicted classes in the hyper-space in between two MNIST images for the classical network.}
\end{figure}
\begin{figure}
    \centering
     \makebox[\textwidth][c]{
     \includegraphics[width=1.0\textwidth]{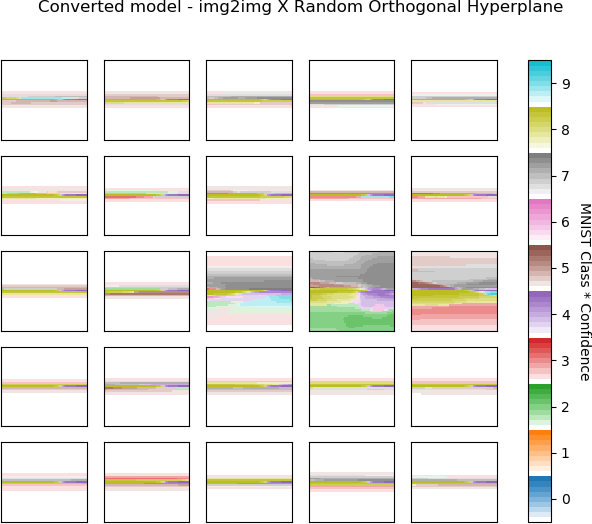}}
    \caption*{Larger version of images in figure \ref{fig:fgsm-img2img}. Boundaries between predicted classes in the hyper-space in between two MNIST images for the converted FGNN network.}
\end{figure}
\begin{figure}
    \centering
     \makebox[\textwidth][c]{
     \includegraphics[width=1.0\textwidth]{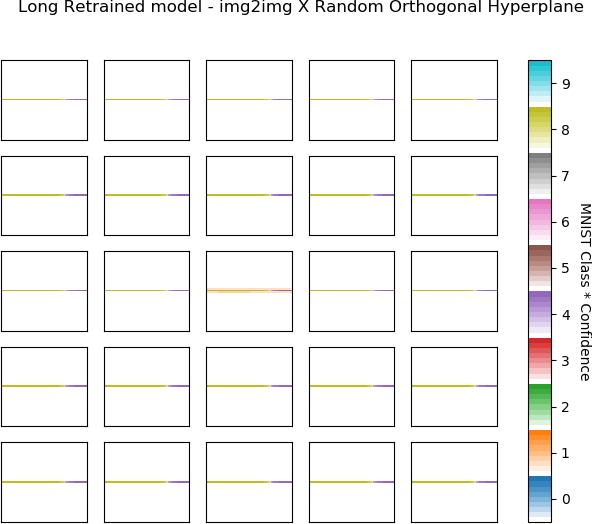}}
    \caption*{Larger version of images in figure \ref{fig:fgsm-img2img}. Boundaries between predicted classes in the hyper-space in between two MNIST images for the long-retrained FGNN network.}
\end{figure}


\bibliography{bibl.bib}

\end{document}